\documentclass[english]{article}
\pdfoutput=1 
\usepackage[T1]{fontenc}
\usepackage[latin9]{inputenc}
\usepackage{babel}
\usepackage{array}
\usepackage{varioref}
\usepackage{float}
\usepackage{amstext}
\usepackage{amssymb}
\usepackage{graphicx}
\usepackage[numbers]{natbib}
\PassOptionsToPackage{normalem}{ulem}
\usepackage{ulem}

\makeatletter

\newcommand{\noun}[1]{\textsc{#1}}
\providecommand{\tabularnewline}{\\}
\newcommand{\lyxdot}{.}

\floatstyle{ruled}
\newfloat{algorithm}{tbp}{loa}
\providecommand{\algorithmname}{Algorithm}
\floatname{algorithm}{\protect\algorithmname}

\usepackage{times}
\usepackage{hyperref}
\usepackage{url}
\usepackage{algorithm,algorithmicx,algpseudocode}

\usepackage{enumitem}
\setlist{nolistsep}
\setlist[itemize]{leftmargin=*}

\makeatother

\begin{document}

\title{Exact gradient updates in time independent of output size for the
spherical loss family}

\author{Pascal Vincent$^{*}$, Alexandre de Brébisson, Xavier Bouthillier\\
Département d'Informatique et de Recherche Opérationnelle\\
Université de Montréal, Montréal, Québec, CANADA\\
$^{*}$and CIFAR}
\maketitle
\begin{abstract}
An important class of problems involves training deep neural networks
with sparse prediction \emph{targets} of very high dimension $D$.
These occur naturally in e.g. neural language models or the learning
of word-embeddings, often posed as predicting the probability of next
words among a vocabulary of size $D$ (e.g. $200\,000$). Computing
the equally large, but typically non-sparse {\normalsize{}$D$}-dimensional
output vector from a last hidden layer of reasonable dimension $d$
(e.g. $500$) incurs a prohibitive $O(Dd)$ computational cost \emph{for
each example}, as does updating the {\normalsize{}$D\times d$} output
weight matrix and computing the gradient needed for backpropagation
to previous layers. While efficient handling of large sparse network
inputs is trivial, the case of large sparse \emph{targets} is not,
and has thus so far been sidestepped with approximate alternatives
such as hierarchical softmax or sampling-based approximations during
training. In this work we develop an original algorithmic approach
which, for a family of loss functions that includes squared error
and spherical softmax, can compute the \emph{exact} loss, gradient
update for the output weights, and gradient for backpropagation, all
in $O(d^{2})$ per example instead of $O(Dd)$, remarkably without
ever computing the $D$-dimensional output. The proposed algorithm
yields a speedup of $\frac{D}{4d}$, i.e. two orders of magnitude
for typical sizes, for that critical part of the computations that
often dominates the training time in this kind of network architecture.
\end{abstract}

\section{Introduction}

Many modern applications of neural networks have to deal with data
represented, or representable, as very large sparse vectors. Such
representations arise in natural language related tasks, where the
dimension $D$ of that vector is typically (a multiple of) the size
of the vocabulary, but also in the sparse user-item matrices of collaborative-filtering
applications. It is trivial to handle very large sparse inputs to
a neural network in a computationally efficient manner: the forward
propagation and update to the input weight matrix after backpropagation
are correspondingly sparse. By contrast, training with very large
sparse prediction \emph{targets} is problematic: even if the target
is sparse, the computation of the equally large network output and
the corresponding gradient update to the huge output weight matrix
are \emph{not sparse} and thus computationally prohibitive. This has
been a practical problem ever since \citet{BenDucVin01} first proposed
using a neural network for learning a language model, in which case
the computed output vector represents the probability of the next
word and is the size of the considered vocabulary, which is becoming
increasingly large in modern applications \citep{collobert:2011b}.
Several approaches have been proposed to attempt to address this difficulty
essentially by sidestepping it. They fall in two categories:
\begin{itemize}
\item \emph{Sampling or selection based approximations} consider and compute
only a tiny fraction of the output's dimensions sampled at random
or heuristically chosen. The reconstruction sampling of \citet{Dauphin2011},
the efficient use of biased importance sampling in \citet{Jean-et-al-ACL2015},
the use of Noise Contrastive Estimation \citep{Gutmann+Hyvarinen-2010}
in \citet{Mnih2013} and \citet{Mikolov-et-al-NIPS2013} all fall
under this category. As does the more recent use of approximate Maximum
Inner Product Search based on Locality Sensitive Hashing techniques\citep{NIPS2014_5329,arxiv1412.7479}
to select a good candidate subset. 
\item \emph{Hierarchical softmax} \citep{Morin+al-2005,Mikolov-et-al-NIPS2013}
imposes a heuristically defined hierarchical tree structure for the
computation of the normalized probability of the target class.
\end{itemize}
Compared to the initial problem of considering all $D$ output dimensions,
both kinds of approaches are crude approximations. In the present
work, we will instead investigate a way to actually perform the \emph{exact}
gradient update that corresponds to considering \emph{all} $D$ outputs,
but do so implicitly, in a computationally efficient manner, without
actually computing the $D$ outputs. This approach works for a relatively
restricted class of loss functions, that we call the \emph{spherical
family}, its simplest member being linear output with squared error
(a natural choice for sparse real-valued regression targets). For
simplicity and clarity we will begin with this squared error case,
presenting the computational challenge that arises in the standard
naive approach in Section \ref{sec:The-problem} and deriving our
algorithmic solution in Section \ref{sec:EfficientAlgoMSE}. We will
then extend our approach to the more general case of loss functions
in the spherical family in Section \ref{sec:generalization-spherical-family}.
In Section \ref{sec:stabilization} we will discuss numerical stability
issues that may arise and detail our numerical stabilization strategy.
Section \ref{sec:experiments} presents experimental validation focusing
on timings obtained with our CPU and GPU implementations of our algorithm
relative to the naive update algorithm.

\section{The problem\label{sec:The-problem}}

\subsection{Problem definition and setup}

We are concerned with gradient-descent based training of a deep feed-forward
neural network with target vectors of very high dimension $D$ (e.g.
$D=200\,000$) but that are sparse, i.e. a comparatively small number,
at most $K\ll D$, of the elements of the target vector are non-zero.
Such a $K$-sparse vector will typically be stored and represented
compactly as $2K$ numbers corresponding to pairs \emph{(index, value)}.
A network to be trained with such targets will naturally have an equally
large output layer of dimension $D$. We can also optionally allow
the input to the network to be a similarly high dimensional sparse
vector of dimension $D_{in}$. Between the large sparse target, output,
and (optionally large sparse) input, we suppose the network's intermediate
hidden layers to be of smaller, more typically manageable, dimension
$d\ll D$ (e.g. $d=500$)\footnote{Our approach does not impose any restriction on the architecture nor
size of the hidden layers, as long as they are amenable to usual gradient
backpropagation.}.

\subsubsection*{\noindent Mathematical notation:}
\begin{itemize}
\item \noindent Vectors are denoted using lower-case letters, e.g. $h$,
and are considered column-vectors; corresponding row vectors are denoted
with a transpose, e.g. $h^{T}$. 
\item \noindent Matrices are denoted using upper-case letters, e.g. $W$,
with $W^{T}$ the transpose of $W$. 
\item \noindent The $j^{th}$ \emph{column} of $W$ is denoted $W_{j}$
, and its $i^{th}$ \emph{row} $W_{i\bullet}$ (both viewed as a column
vector). 
\item \noindent $U^{-T}=\left(U^{-1}\right)^{T}$ denotes the transpose
of the inverse of a square matrix. 
\item \noindent $\mathbf{1}_{D}$ denotes a $D$-dimensional column vector
filled with ones. 
\item \noindent $\mathbf{1}_{i\in\mathcal{A}(y)}$ denotes an indicator
function whose value will be 1 if $i\in\mathcal{A}(y)$ and 0 otherwise.
\item \noindent $\mathrm{onehot}_{D}(j)=\left\{ \mathbf{1}_{i=j}\right\} _{i=1}^{D}$
is the $D$-dimensional column vector filled with zeros except at
index $j$ where its value is 1.
\item \noindent $\mathbf{I}_{d}$ is the $d\times d$ identity matrix.
\end{itemize}

\subsubsection*{\noindent Network architecture}

\begin{figure}
\centering{}\includegraphics[width=0.8\columnwidth]{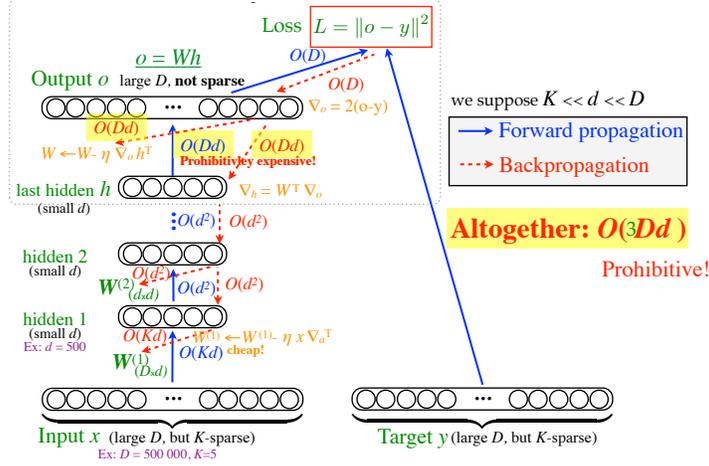}\protect\caption{\label{fig:network}The computational problem posed by very large
sparse targets. Dealing with sparse input efficiently is trivial,
with both the forward and backward propagation phases easily achieved
in $O(Kd)$. However this is not the case with large sparse targets.
They incur a prohibitive computational cost of $O(Dd)$ at the output
layer as forward propagation, gradient backpropagation and weight
update each require accessing all $D\times d$ elements of the large
output weight matrix.}
\end{figure}

\noindent We consider a standard feed forward neural network architecture
as depicted in Figure \ref{fig:network}. An input vector $x\in\mathbb{R}^{D_{in}}$
is linearly transformed into a linear activation $a^{(1)}=W^{(1)T}x+b^{(1)}$
through a $D_{in}\times d$ input weight matrix $W^{(1)}$ (and an
optional bias vector $b^{(1)}\in\mathbb{R}^{d}$). This is typically
followed by a non-linear transformation $s$ to yield the representation
of the first hidden layer $h^{(1)}=s(a^{(1)})$. This first hidden
layer representation is then similarly transformed through a number
of subsequent non-linear layers (that can be of any usual kind amenable
to backpropagation) e.g. $h^{(k)}=s(a^{(k)})$ with $a^{(k)}=W^{(k)T}h^{(k-1)}+b^{(k)}$
until we obtain last hidden layer representation $h=h^{(m)}$. We
then obtain the final $D$-dimensional network output as $o=Wh$ where
$W$ is a $D\times d$ output weight matrix, which will be our main
focus in this work. Finally, the network's $D$-dimensional output
$o$ is compared to the $D$-dimensional target vector $y$ associated
with input $x$ using squared error, yielding loss $L=\|o-y\|^{2}$.

\subsubsection*{\noindent Training procedure }

\noindent This architecture is a typical (possibly deep) multi-layer
feed forward neural network architecture with a \emph{linear output
layer} and \emph{squared error loss}. Its parameters (weight matrices
and bias vectors) will be trained by gradient descent, using gradient
backpropagation \citet{Rumelhart86b-small,LeCun85,LeCun86} to efficiently
compute the gradients. The procedure is shown in Figure \ref{fig:network}.
Given an example from the training set as an \emph{(input,target)}
pair $(x,y)$, a pass of forward propagation proceeds as outlined
above, computing the hidden representation of each hidden layer in
turn based on the previous one, and finally the network's predicted
output $o$ and associated loss $L$. A pass of gradient backpropagation
then works in the opposite direction, starting from $\nabla_{o}=\frac{\partial L}{\partial o}=2(o-y)$
and propagating back the gradients $\nabla_{h^{(k)}}=\frac{\partial L}{\partial h^{(k)}}$
and $\nabla_{a^{(k)}}=\frac{\partial L}{\partial a^{(k)}}$ upstream
through the network. The corresponding gradient contributions on parameters
(weights and biases), collected along the way, are straightforward
once we have the associated $\nabla_{a^{(k)}}$. Specifically they
are $\nabla_{b^{(k)}}=\nabla_{a^{(k)}}$ and $\nabla_{W^{(k)}}=h^{(k-1)}(\nabla_{a^{(k)}})^{T}$.
Similarly for the input layer $\nabla_{W^{(1)}}=x(\nabla_{a^{(1)}})^{T}$,
and for the output layer $\nabla_{W}=(o-y)h^{T}$ . Parameters are
then updated through a gradient descent step $W^{(k)}\leftarrow W^{(k)}-\eta\nabla_{W^{(k)}}$
and $b^{(k)}\leftarrow b^{(k)}-\eta\nabla_{b^{(k)}}$, where $\eta$
is a positive learning-rate. Similarly for the output layer which
will be our main focus here: $W\leftarrow W-\eta\nabla_{W}$.

\subsection{\label{sub:easy-part}The easy part: input layer forward propagation
and weight update}

It is easy and straightforward to efficiently compute the forward
propagation, and the backpropagation and weight update part for the
\emph{input layer} when we have a very large $D_{in}$-dimensional
but $K-$sparse input vector $x$ with appropriate sparse representation.
Specifically we suppose that $x$ is represented as a pair of vectors
$u,v$ of length (at most) $K$, where $u$ contains integer indexes
and $v$ the associated real values of the elements of $x$ such that
$x_{i}=0$ if $i\notin u$, and $x_{u_{k}}=v_{k}$.
\begin{itemize}
\item \noindent \textbf{Forward propagation through the input layer:} The
sparse representation of $x$ as the positions of $K$ elements together
with their value makes it cheap to compute $W^{(1)T}x$. Even though
$W^{(1)}$ may be a huge full $D_{in}\times d$ matrix, only $K$
of its rows (those corresponding to the non-zero entries of $x$)
need to be visited and summed to compute $W^{(1)T}x$. Precisely,
with our $(u,v)$ sparse representation of $x$ this operation can
be written as$W^{(1)T}x=\sum_{k=1}^{K}v_{k}W_{:u_{k}}^{(1)}$where
each $W_{:u_{k}}^{(1)}$ is a $d$-dimensional vector, making this
an $O(Kd)$ operation rather than $O(Dd)$.
\item \textbf{Gradient and update through input layer:} Let us for now suppose
that we were able to get gradients (through backpropagation) up to
the first hidden layer activations $a^{(1)}\in\mathbb{R}^{d}$ in
the form of gradient vector $\nabla_{a^{(1)}}=\frac{\partial L}{\partial a^{(1)}}$.
The corresponding gradient-based update to input layer weights $W^{(1)}$
is simply $W^{(1)}\leftarrow W^{(1)}-\eta x(\nabla_{a^{(1)}})^{T}$.
This is a rank-one update to $W^{(1)}$. Here again, we see that only
the $K$ rows of $W^{(1)}$ associated to the (at most) $K$ non-zero
entries of $x$ need to be modified. Precisely this operation can
be written as:$W_{:u_{k}}^{(1)}\leftarrow W_{:u_{k}}^{(1)}-\eta v_{k}\nabla_{a^{(1)}}\,\,\,\forall k\in\{1,\ldots,K\}$
making this again a $O(Kd)$ operation rather than $O(Dd)$.
\end{itemize}

\subsection{\label{sub:hard-part}The hard part: output layer propagation and
weight update}

Given some network input $x$ we suppose we can compute without difficulty
through forward propagation the associated last hidden layer representation
$h\in\mathbb{R}^{d}$. From then on:
\begin{itemize}
\item Computing the final output $o=Wh$ incurs a prohibitive computational
cost of $O(Dd)$ since $W$ is a full $D\times d$ matrix. Note that
there is a-priori no reason for representation $h$ to be sparse (e.g.
with a sigmoid non-linearity) but even if it was, this would not fundamentally
change the problem since it is $D$ that is extremely large, and we
supposed $d$ reasonably sized already. Computing the residual $(\mbox{o}-y)$
and associated squared error loss $\|\mbox{o}-y\|^{2}$ incurs an
additional $O(D)$ cost.
\item The gradient on $h$ that we need to backpropagate to lower layers
is $\nabla_{h}=\frac{\partial L}{\partial h}=2W^{T}(o-y)$ which is
another $O(Dd)$ matrix-vector product.
\item Finally, when performing the corresponding output weight update $W\leftarrow W-\eta(o-y)h^{T}$
we see that it is a rank-one update that updates all $D\times d$
elements of $W$, which again incurs a prohibitive $O(Dd)$ computational
cost.
\end{itemize}
For very large $D$, all these three $O(Dd)$ operations are prohibitive,
and the fact that $y$ is sparse, seen from this perspective, doesn't
help, since neither $o$ nor $o-y$ will be sparse.

\section{A computationally efficient algorithm for performing the exact online
gradient update\label{sec:EfficientAlgoMSE}}

Previously proposed workarounds are approximate or use stochastic
sampling. We propose a different approach that results in the \emph{exact
same}, yet efficient gradient update, remarkably without ever having
to compute large output $o$.

\subsection{\label{sub:efficient-squared-loss}Computing the squared error loss
$L$ and the gradient with respect to $h$ efficiently}

Suppose that, we have, for a network input example $x$, computed
the last hidden representation $h\in\mathbb{R}^{d}$ through forward
propagation. The network's $D$ dimensional output $o=Wh$ is then
in principle compared to the high dimensional target $y\in\mathbb{R}^{D}$.
The corresponding squared error loss is $L=\left\Vert Wh-y\right\Vert ^{2}$.
As we saw in Section \ref{sub:hard-part}, computing it in the direct
naive way would have a prohibitive computational complexity of $O(Dd+D)=O(Dd)$
because computing output $Wh$ with a full $D\times d$ matrix $W$
and a typically non-sparse $h$ is $O(Dd)$. Similarly, to backpropagate
the gradient through the network, we need to compute the gradient
of loss $L$ with respect to last hidden layer representation $h$.
This is $\nabla_{h}=\frac{\partial L}{\partial h}=\frac{\partial\left\Vert Wh-y\right\Vert ^{2}}{\partial h}=2W^{T}(Wh-y)$.
So again, if we were to compute it directly in this manner, the computational
complexity would be a prohibitive $O(Dd)$. \textbf{Provided we have
maintained an up-to-date matrix $Q=W^{T}W$}, which is of reasonable
size $d\times d$ and can be cheaply maintained as we will see in
Section \ref{sub:bookkeeping}, we can rewrite these two operations
so as to perform them in $O(d^{2})$:

\paragraph*{Loss computation:}

\begin{eqnarray}
L & = & \|\overbrace{Wh}^{O(Dd)}-y\|^{2}\nonumber \\
 & = & \left(Wh-y\right)^{T}\left(Wh-y\right)\nonumber \\
 & = & h^{T}W^{T}Wh-y^{T}Wh-h^{T}W^{T}y+y^{T}y\nonumber \\
 & = & h^{T}Qh-2h^{T}(W^{T}y)+y^{T}y\nonumber \\
 & = & h^{T}(\underbrace{Qh}_{O(d^{2})}-2\underbrace{W^{T}y}_{O(Kd)})+\underbrace{y^{T}y}_{O(K)}\label{eq:L}
\end{eqnarray}

\paragraph*{Gradient on $h$:}

\begin{eqnarray}
\nabla_{h}=\frac{\partial L}{\partial h} & = & \frac{\partial\|Wh-y\|^{2}}{\partial h}\nonumber \\
 & = & 2W^{T}(Wh-y)\nonumber \\
 & = & 2\left(W^{T}Wh-W^{T}y\right)\nonumber \\
 & = & 2(\underbrace{Qh}_{O(d^{2})}-\underbrace{W^{T}y}_{O(Kd)})\label{eq:gradH}
\end{eqnarray}

The terms in $O(Kd)$ and $O(K)$ are due to leveraging the $K$-sparse
representation of target vector $y$. With $K\ll D$ and $d\ll D$,
we get altogether a computational cost of $O(d^{2})$ which can be
several orders of magnitude cheaper than the prohibitive $O(Dd)$
of the direct approach.

\subsection{Efficient gradient update of $W$}

The gradient of the squared error loss with respect to output layer
weight matrix $W$ is $\frac{\partial L}{\partial W}=\frac{\partial\left\Vert Wh-y\right\Vert ^{2}}{\partial W}=2(Wh-y)h^{T}$.
And the corresponding gradient descent update to $W$ would be $W_{new}\leftarrow W-2\eta(Wh-y)h^{T}$,
where $\eta$ is a positive learning rate. Again, computed in this
manner, this induces a prohibitive $O(Dd)$ computational complexity,
both to compute output and residual $Wh-y$, and then to update all
the $Dd$ elements of $W$ (since generally neither $Wh-y$ nor $h$
will be sparse). All $D\times d$ elements of $W$ must be accessed
during this update. On the surface this seems hopeless. But we will
now see how we can achieve the \emph{exact} same update on $W$ in
$O(d^{2})$. The trick is to represent $W$ \emph{implicitly} as the
factorization\footnote{Note that we never \emph{factorize} a pre-exisitng arbitrary $W$,
which would be prohibitive as $W$ is huge. We will no longer store
a $W$ nor work on it explicitly, but only matrices $V$ and $U$
which implicitly represent $W$.} $\underbrace{W}_{D\times d}=\underbrace{V}_{D\times d}\underbrace{U}_{d\times d}$and
update $U$ and $V$ instead:

\begin{eqnarray}
\mathbf{a)}\,\,U_{new} & = & U-2\eta(Uh)h^{T}\label{eq:updateU}\\
\mathbf{b)}\,\,V_{new} & = & V+2\eta y(U_{new}^{-T}h){}^{T}\label{eq:updateV}
\end{eqnarray}

This results in \emph{implicitly} updating $W$ as we did \emph{explicitly}
in the naive approach as we now prove:

\begin{eqnarray*}
V_{new}U_{new} & = & (V+2\eta y(U_{new}^{-T}h){}^{T})\,U_{new}\\
 & = & VU_{new}+2\eta y(U_{new}^{-T}h){}^{T}U_{new}\\
 & = & VU_{new}+2\eta yh^{T}U_{new}^{-1}U_{new}\\
 & = & V(U-2\eta(Uh)h^{T})+2\eta yh^{T}(U_{new}^{-1}U_{new})\\
 & = & VU-2\eta VUhh^{T}+2\eta yh^{T}\\
 & = & VU-2\eta(VUh-y)h^{T}\\
 & = & W-2\eta(Wh-y)^{T}h^{T}\\
 & = & W_{new}
\end{eqnarray*}

We see that the update of $U$ in Eq.~\ref{eq:updateU} is a simple
$O(d^{2})$ operation. Following this simple rank-one update to $U$,
we can use the Sherman-Morrison formula to derive the corresponding
rank-one update to $U^{-T}$ which will also be $O(d^{2})$:

\begin{eqnarray}
U_{new}^{-T} & = & U^{-T}+\frac{2\eta}{1-2\eta\left\Vert h\right\Vert ^{2}}(U^{-T}h)h^{T}\label{eq:updateU-T}
\end{eqnarray}
It is then easy to compute the $U_{new}^{-T}h$, an $O(d^{2})$ operation
needed in Eq.~\ref{eq:updateV}. The ensuing rank-one update of $V$
in Eq~\ref{eq:updateV}, thanks to the $K$-sparsity of $y$ is only
$O(Kd)$: only the$K$ rows $V$ associated to non-zero elements in
$y$ are accessed and updated, sited of \emph{all} $D$ rows of $W$
we had to modify in the naive update!

\subsection{Adapting the computation of $L$ and $\nabla_{h}$ to the factored
representation of $W$}

With the factored representation of $W$ as $VU$, we only have $W$
implicitly, so the $W^{T}y$ terms that entered in the computation
of $L$ and $\nabla_{h}$ in the previous section (Eq. \vref{eq:L}
and \vref{eq:gradH}) need to be adapted slightly as $\hat{y}=W^{T}y=U^{T}(V^{T}y)$,
which becomes $O(d^{2}+Kd)$ rather than $O(Kd)$ in computational
complexity. But this doesn't change the overall $O(d^{2})$ complexity
of these computations.

The adapted update computation of $L$ and $\nabla_{h}$ can thus
be expressed simply as:

\begin{equation}
\nabla_{h}=2\underbrace{(\underbrace{Qh}_{\hat{h}}-\underbrace{U^{T}(V^{T}y)}_{\hat{y}})}_{\hat{z}}\label{eq:online-grad-h-fact}
\end{equation}

and 
\begin{equation}
L=h^{T}(\underbrace{Qh}_{\hat{h}}-2\underbrace{U^{T}(V^{T}y)}_{\hat{y}})+y^{T}y\label{eq:online-L-fact}
\end{equation}

\subsection{\label{sub:bookkeeping}Bookkeeping: keeping an up-to-date $Q$ and
$U^{-T}$}

We have already seen, in Eq.~\ref{eq:updateU-T}, how we can cheaply
maintain an up-to-date $U^{-T}$ following our update of $U$. Similarly,
following our updates to $U$ and $V$, we need to keep an up-to-date
$Q=W^{T}W$ which is needed to efficiently compute the loss $L$ (Eq.~\ref{eq:L})
and gradient $\nabla_{h}$ (Eq.~\ref{eq:gradH}). We have shown that
updates to $U$ and $V$ in equations~\ref{eq:updateU} and \ref{eq:updateV}
are equivalent to implicitly updating $W$ as $W_{new}\leftarrow W-2\eta(Wh-y)h^{T}$,
and this translates into the following update to $Q=W^{T}W$:

\begin{eqnarray}
Q_{new} & = & Q-\eta\left(h\nabla_{h}^{T}+\nabla_{h}h^{T}\right)+(4\eta^{2}L)hh^{T}\label{eq:updateQ}
\end{eqnarray}

One can see that this last bookkeeping operation also has a $O(d^{2})$
computational complexity.

\subsubsection*{Proof that this update to $Q$ corresponds to the update \textmd{$W_{new}\leftarrow2(Wh-y)h^{T}$}}

\begin{eqnarray*}
W_{new}^{T}W_{new} & = & \left(W-2\eta(Wh-y)h^{T}\right)^{T}\left(W-2\eta(Wh-y)h^{T}\right)\\
W_{new}^{T}W_{new} & = & W^{T}W-2\eta h(Wh-y)^{T}W-2\eta W^{T}(Wh-y)h^{T}\\
 &  & +4\eta^{2}h(Wh-y)^{T}(Wh-y)h^{T}\\
W_{new}^{T}W_{new} & = & Q-2\eta\left(hh^{T}W^{T}W-hy^{T}W\right)-2\eta\left(W^{T}Whh^{T}-W^{T}yh^{T}\right)\\
 &  & +4\eta^{2}h(h^{T}W^{T}Wh-h^{T}W^{T}y-y^{T}Wh+y^{T}y)h^{T}\\
W_{new}^{T}W_{new} & = & Q-2\eta\left(hh^{T}Q-h(W^{T}y)^{T}\right)-2\eta\left(Qhh^{T}-(W^{T}y)h^{T}\right)\\
 &  & +4\eta^{2}h(h^{T}Qh-h^{T}(W^{T}y)-(W^{T}y)^{T}h+y^{T}y)h^{T}\\
W_{new}^{T}W_{new} & = & Q-2\eta h\left(h^{T}Q-(W^{T}y)^{T}\right)-2\eta\left(Qh-W^{T}y\right)h^{T}\\
 &  & +4\eta^{2}h(h^{T}Qh-2h^{T}W^{T}y+y^{T}y)h^{T}\\
\\
W_{new}^{T}W_{new} & = & Q-\eta h(\underbrace{2(Qh-W^{T}y)}_{\nabla_{h}})^{T}-\eta(\underbrace{2(Qh-W^{T}y)}_{\nabla_{h}})h^{T}\\
 &  & +4\eta^{2}h\underbrace{(h^{T}(Qh-2W^{T}y)+y^{T}y)}_{L}h^{T}
\end{eqnarray*}

where we see that the last term uses the expression of $L$ from Eq.
\vref{eq:L} and the first two terms uses the expression of $\nabla_{h}$
from Eq. \ref{eq:online-grad-h-fact}: $\nabla_{h}=2(Qh-U^{T}(V^{T}y))=2(Qh-W^{T}y)$.
Thus we have shown that

\begin{eqnarray*}
W_{new}^{T}W_{new} & = & Q-\eta h\nabla_{h}^{T}-2\eta\nabla_{h}h^{T}+4\eta^{2}hLh^{T}\\
 & = & Q-\eta\left(h\nabla_{h}^{T}+\nabla_{h}h^{T}\right)+(4\eta^{2}L)hh^{T}
\end{eqnarray*}

which is the update $Q_{new}$ that we gave in Eq. \ref{eq:updateQ}
above.

\subsubsection*{}

\subsection{Putting it all together: detailed online update algorithm and expected
benefits\label{sub:online-altogether}}

We have seen that we can efficiently compute cost $L$, gradient with
respect to $h$ (to be later backpropagated further) as well as updating
$U$ and $V$ and performing the bookkeeping for $U^{-T}$ and $Q$.
Here we put everything together. The parameters of the output layer
that we will learn are $V,U$ and implicitly represent $W$ as $W=VU$.
We first need to initialize these parameter matrices, as well as bookkeeping
matrices $Q$ and $U^{-T}$ in a consistent way, as explained in Algo.
\ref{alg:mse-initialization}. We then iterate over the following:
\begin{itemize}
\item pick a next \emph{input,target} example $x,y$ (where $y$ is $K$-sparse
and uses an appropriate sparse representation)
\item perform forward propagation through all layers of the network up to
the last hidden layer, to compute last hidden layer representation
$h=h(x)$, that should include a constant 1 first element.
\item execute Algo. \ref{alg:online-mse}, that we put together from the
equations derived above, and that will: compute the associated squared
error loss $L$, perform an implicit gradient update step on $W$
by correspondingly updating $V$ and $U$ in a computationally efficient
manner, update bookkeeping matrices $Q$ and $U^{-T}$ accordingly,
and compute and return the gradient of the loss with respect to the
last hidden layer $\nabla_{h}$ 
\item having $\nabla_{h}$, further backpropagate the gradients upstream,
and use them to update the parameters of all other layers 
\end{itemize}
Having $K\ll d\ll D$ we see that the update algorithm we developed
requires $O(d^{2})$ operations, whereas the standard approach required
$O(Dd)$ operations. If we take $K\approx d$ , we may state more
precisely that the proposed algorithm, for computing the loss and
the gradient updates will require roughly $12d^{2}$ operations whereas
the standard approach required roughly $3Dd$ operations. So overall
the proposed algorithm change corresponds to a computational speedup
by a factor of $\frac{D}{4d}$. For $D=200\,000$ and $d=500$ the
expected speedup is thus \textbf{100}. Note that the advantage is
not only in \emph{computational} complexity, but also in \emph{memory
access}. For each example, the standard approach needs to access and
change all $D\times d$ elements of matrix $W$, whereas the proposed
approach only accesses the much smaller number $K\times d$ elements
of $V$ as well as the three $d\times d$ matrices $U$, $U^{-T}$,
and $Q$. So overall we have a \textbf{substantially faster algorithm
whose complexity is independent of $D$}, which, while doing so \emph{implicitly},
will nevertheless perform the \emph{exact same} gradient update as
the standard $O(Dd)$ approach. We want to emphasize here that this
approach is entirely different from simply chaining 2 linear layers
$U$ and $V$ and performing ordinary gradient descent updates on
these: this would result in the same prohibitive computational complexity
as the standard approach, and such ordinary separate gradient updates
to $U$ and $V$ would \emph{not} be equivalent to the ordinary gradient
update to $W=VU$.

\begin{algorithm}
\begin{centering}
\protect\caption{\label{alg:mse-initialization}Initialization of output layer parameters
$V,U$ and bookkeeping matrices $Q,U^{-T}$}

\par\end{centering}
\begin{itemize}
\item we can initialize $D\times d$ matrix $V$ randomly as we would have
initialized $W$ so that we initially have $V=W$.\\
Alternatively we can initialize $V$ to 0 (there won't be symmetry
breaking issues with having $W$ initially be 0 provided the other
layers are initialized randomly, since varying inputs and targets
will naturally break symmetry for the output layer)
\item initialize $Q\leftarrow V^{T}V$ (or more cheaply initialize $Q\leftarrow0$
if we have initialized $V$ to 0).
\item we initialize $U$ to the identity: $U\leftarrow\mathbf{I}_{d}$ so
that, trivially, we initially have $VU=W$.
\item initialize $U^{-T}\leftarrow\mathbf{I}_{d}$\end{itemize}
\end{algorithm}

\begin{algorithm}
\begin{centering}
\protect\caption{\label{alg:online-mse}Efficient computation of cost $L$, gradient
$\nabla h$, and update to parameters $U$ and $V$ for squared error,
in the online case}

\par\end{centering}

\textbf{Inputs} (besides above parameters $V,U,Q,U^{-T}$): 
\begin{itemize}
\item $h\in\mathbb{R}^{d}$ hidden representation vector for one example
$h\in\mathbb{R}^{d}$ 
\item $y\in\mathbb{R}^{D}$ associated \emph{\uline{K}}\uline{-sparse}
target vector stored using a sparse representation (indices and values
of non-zero elements)
\item $\eta\in\mathbb{R}^{+}$learning rate for the update
\end{itemize}
\textbf{Outputs:} 
\begin{itemize}
\item $L\in\mathbb{R}$ the squared error loss for this example
\item updated parameters and bookkeeping matrices $U_{new},V_{new},Q_{new},U_{new}^{-T}$
\item $\nabla_{h}\in\mathbb{R}^{d}$ the gradient of the loss with respect
to $h$, to further backpropagate upstream.
\end{itemize}
\textbf{Algorithm:}

\begin{centering}
\begin{tabular}{|>{\raggedright}p{0.05\textwidth}|>{\raggedright}p{0.3\textwidth}|>{\centering}p{0.15\textwidth}|>{\centering}p{0.15\textwidth}|}
\hline 
\textbf{Step \#} & \textbf{Operation} & \textbf{Computational complexity} & \textbf{Approximate number of elementary operations (multiply-adds)}\tabularnewline
\hline 
\hline 
1: & $\hat{h}=Qh$  & $O(d^{2})$ & $d^{2}$\tabularnewline
\hline 
2: & $\hat{y}=U^{T}(V^{T}y)$ & $O(Kd+d^{2})$ & $Kd+d^{2}$\tabularnewline
\hline 
3: & $\nabla_{h}=2(\hat{h}-\hat{y})$ & $O(d)$ & $d$\tabularnewline
\hline 
4: & $L=h^{T}\hat{h}-2h^{T}\hat{y}+y^{T}y$ & $O(2d+K)$ & $2d+K+1$\tabularnewline
\hline 
5: & $U\leftarrow U-2\eta(Uh)h^{T}$ & $O(d^{2})$ & $2d^{2}+d$\tabularnewline
\hline 
6: & $U^{-T}\leftarrow U^{-T}+\frac{2\eta}{1-2\eta\left\Vert h\right\Vert ^{2}}(U^{-T}h)h^{T}$

{[} from Sherman-Morrison formula {]} & $O(d^{2})$ & $2d^{2}+2d+3$\tabularnewline
\hline 
7: & $V\leftarrow V+2\eta y(U^{-T}h){}^{T}$ where we must use the freshly
updated $U^{-T}$ resulting from step 6) & $O(d^{2}+Kd)$ & $d^{2}+K+Kd$\tabularnewline
\hline 
8: & $Q\leftarrow Q-\eta\left(h\nabla_{h}^{T}+\nabla_{h}h^{T}\right)+(4\eta^{2}L)hh^{T}$ & $O(d^{2})$ & $4+2d+3d^{2}$\tabularnewline
\hline 
 & \textbf{Altogether:} & $O(d^{2})$

provided $K<d\ll D$ & $\approx12d^{2}$ elementary operations\tabularnewline
\hline 
\end{tabular}
\par\end{centering}

\end{algorithm}

\subsection{Minibatch version of the algorithm for squared error}

The algorithm we derived for online gradient is relatively straightforward
to extend to the case of minibatches containing $m$ examples. We
iniialize parameters as in the online case follpwing Algo. \ref{alg:mse-initialization}
and apply the same training procedure outlined in Section. \ref{sub:online-altogether},
but now using minibatches containing $m$ examples, rather than a
single example vector. The corresponding update and gradient computation
is given in Algorithm~ \ref{alg:minibatch-mse} which follows equivalent
steps to the online version of Algorithm \ref{alg:online-mse}, but
using matrices with $m$ columns in place of single column vectors.
For example step 3 which in the online algorithm was $\nabla_{h}=2(\hat{h}-\hat{y})$
using $d-$dimensional vectors becomes in the minibatch version $\nabla_{H}=2(\hat{H}-\hat{Y})$
using $d\times m$ matrices instead.

Note that in the minibatch version, in step 6, we update $U^{-T}$
based on the Woodbury equation, which generalizes the Sheman-Morrison
formula for $m>1$ and involves inverting an $m\times m$ matrix,
an $O(m^{3})$ operation. But depending on the size of the minibatch
$m$, it may become more efficient to solve the corresponding linear
equations for each minibatch from scratch every time, rather than
inverting that $m\times m$ matrix. In which case we won't need to
maintain an $U^{-T}$ at all. Or in cases of minibatches containing
more than $d$ examples, it may even become more efficient to invert
$U$ from scratch every time. 

In step 9, the update $Q_{new}$ for $Q$ corresponds to the implicit
weight update $W_{new}\leftarrow W-2\eta(WH-Y)H^{T}$ as we now prove: 

We will use the following precomputed quantities: $Q=W^{T}W$, $\hat{H}=QH$
and $\hat{Y}=W^{T}Y=U^{T}(V^{T}Y)$ and $\nabla_{H}=2(\hat{H}-\hat{Y})$.

\begin{eqnarray*}
Q_{new} & = & W_{new}^{T}W_{new}\\
 & = & \left(W-2\eta(WH-Y)H^{T}\right)^{T}\left(W-2\eta(WH-Y)H^{T}\right)\\
 & = & W^{T}W-2\eta H(WH-Y)^{T}W-2\eta W^{T}(WH-Y)H^{T}\\
 &  & +4\eta^{2}H(WH-Y)^{T}(WH-Y)H^{T}\\
 & = & Q-2\eta\left(HH^{T}W^{T}W-HY^{T}W\right)-2\eta\left(W^{T}WHH^{T}-W^{T}YH^{T}\right)\\
 &  & +4\eta^{2}H(H^{T}W^{T}WH-H^{T}W^{T}Y-Y^{T}WH+Y^{T}Y)H^{T}\\
 & = & Q-2\eta\left(HH^{T}Q-H(W^{T}Y)^{T}\right)-2\eta\left(QHH^{T}-(W^{T}Y)H^{T}\right)\\
 &  & +4\eta^{2}H(H^{T}QH-H^{T}(W^{T}Y)-(W^{T}Y)^{T}H+Y^{T}Y)H^{T}\\
 & = & Q-2\eta\left(H\hat{H}^{T}-H\hat{Y}^{T}+\hat{H}H^{T}-\hat{Y}H^{T}\right)\\
 &  & +4\eta^{2}H(H^{T}\hat{H}-H^{T}\hat{Y}-\hat{Y}^{T}H+Y^{T}Y)H^{T}\\
 & = & Q-2\eta\left(H(\hat{H}-\hat{Y})^{T}+(\hat{H}-\hat{Y})H^{T}\right)+4\eta^{2}H(H^{T}(\hat{H}-\hat{Y})-\hat{Y}^{T}H+Y^{T}Y)H^{T}\\
 & = & Q-\eta\left(H(2(\hat{H}-\hat{Y}))^{T}+(2(\hat{H}-\hat{Y}))H^{T}\right)+4\eta^{2}H(H^{T}(\hat{H}-\hat{Y})-\hat{Y}^{T}H+Y^{T}Y)H^{T}\\
 & = & Q-\eta\left(H\nabla_{H}^{T}+\nabla_{H}H^{T}\right)+4\eta^{2}H\underbrace{\left(H^{T}\hat{Z}-\hat{Y}^{T}H+Y^{T}Y\right)}_{M}H^{T}
\end{eqnarray*}

which is the update of $Q$ we use in in step 8 of Algorithm \vpageref{alg:online-mse}.

\begin{algorithm}[H]
\protect\caption{\label{alg:minibatch-mse}Minibatch version of the update algorithm
for squared error}

\textbf{Inputs} (besides above parameters $V,U,Q,U^{-T}$): 
\begin{itemize}
\item parameters and bookkeeping matrices: $U,\text{ }V,\text{ }Q,\text{ }U^{-T}$
\item $H$ : a $d\times m$ matrix whose $m$ \uline{columns} contain
the last hidden layer representation vectors for $m$ example (with
an appended constant 1 element to account for an output bias). 
\item $Y$ : a $D\times m$ sparse target matrix. Each of its $m$ columns
is the $K$-sparse target vector associated to one example of the
minibatch, stored using a sparse representation (indices and values
of non-zero elements).
\item $\eta\in\mathbb{R}^{+}$learning rate for the update
\end{itemize}
\textbf{Updates:} 
\begin{itemize}
\item parameters and bookkeeping matrices: $U,\text{ }V,\text{ }Q,\text{ }U^{-T}$
\end{itemize}
\textbf{Outputs:} 
\begin{itemize}
\item $L\in\mathbb{R}$ the sum of squared error losses for the $m$ examples
of the minibatch
\item $\nabla_{H}$ a $d\times m$ matrix whose $m$ columns contain the
gradient of the loss with respect to $H$, to further backpropagate
upstream.
\end{itemize}
\textbf{Algorithm:}

\begin{tabular}{|>{\raggedright}p{0.05\textwidth}|>{\raggedright}p{0.4\textwidth}|>{\centering}p{0.15\textwidth}|>{\raggedright}p{0.3\textwidth}|}
\hline 
\textbf{Step \#} & \textbf{Operation} & \textbf{Computation complexity} & \textbf{Approximate number of elementary operations (multiply-adds)}\tabularnewline
\hline 
\hline 
1: & $\hat{H}=QH$  & $O(md^{2})$ & $md^{2}$\tabularnewline
\hline 
2: & $\hat{Y}=U^{T}(V^{T}Y)$ & $O(mKd+md^{2})$ & $mKd+md^{2}$\tabularnewline
\hline 
3: & $\nabla_{H}=2(\hat{H}-\hat{Y})$ & $O(md)$ & $md$\tabularnewline
\hline 
4a: & $M=H^{T}\hat{H}-(\hat{Y}^{T}H+H^{T}\hat{Y})+Y^{T}Y$ 

 & $O(m^{2}d+m^{2}K)$ & $2m^{2}d+m^{2}K$\tabularnewline
\hline 
4b: & $L=\mathrm{Tr}(M)$ & $O(m)$ & $m$\tabularnewline
\hline 
5: & $U\leftarrow U-2\eta(UH)H^{T}$ & $O(md^{2})$ & $2md^{2}+md$\tabularnewline
\hline 
6: & $U^{-T}\leftarrow U^{-T}-(U^{-T}H)\left((H^{T}H-\frac{1}{2\eta}\mathbf{I}_{m})^{-1}H^{T}\right)$ 

{[} from Woodbury identity {]} & $O(m^{2}d+m^{3}+md^{2})$ & $2md^{2}+m+\frac{2}{3}m^{3}+m^{2}d$ (we count $\frac{2}{3}m^{3}$
operations for inversion of a $m\times m$ matrix) \tabularnewline
\hline 
7: & $V\leftarrow V+2\eta Y(U^{-T}H){}^{T}$ where we must use the freshly
updated $U^{-T}$ resulting from step 6) & $O(md^{2}+mKd)$ & $md^{2}+mK+mKd$\tabularnewline
\hline 
8: & $Q\leftarrow Q-\eta\left(H\nabla_{H}^{T}+\nabla_{H}H^{T}\right)+4\eta^{2}(HM)H^{T}$ & $O(md^{2}+m^{2}d)$ & $m^{2}d+3md^{2}+2d^{2}$\tabularnewline
\hline 
 & \textbf{Altogether:} & $O(md^{2})$

provided $K<m<d\ll D$. & $\approx10md^{2}+3m^{2}d+m^{3}$ elementary operations when $K=1$\tabularnewline
\hline 
\end{tabular}

Note that if we chose $m>d$ we will not perform step 7 based on the
Woodbury identity, which would be wasteful, but instead directly recompute
the inverse of $U_{new}$ in $O(d^{3})$. The overall complexity remains
$O(md^{2})$ in this case also.
\end{algorithm}

\section{Generalizing to a broader family of loss functions}

\label{sec:generalization-spherical-family}

Let $o=Wh$ the linear activations computed at the output layer. The
approach that we detailed for linear output and squared error can
be extended to a more general family of loss functions: basically
any loss function $\ell$ that can be expressed using only the $o_{c}$
associated to non-zero $y_{c}$ together with $q=\|o\|^{2}=\sum_{j}o_{j}^{2}$
the squared norm of the whole output vector, and optionally $s=\mathrm{sum}(o)=\sum_{j}o_{j}$
which we will see that we can both compute cheaply. We call this family
of loss functions the \emph{spherical family of loss functions} or
in short \emph{spherical losses,} defined more formally as the family
of losses that can be expressed as:

\textbf{}

\[
L=\mathcal{\ell}(~\|o\|^{2},~\mathrm{sum}(o),~\mathcal{K},~o_{\mathcal{K}},~y_{\mathcal{K}})
\]

where $\mathcal{K}$ denotes the vector of indices of $y$ of cardinality
at most $K\ll D$ that is associated to non-zero elements of $y$
in a sparse representation of$y$; $y_{\mathcal{K}}$ is the corresponding
vector of values of $y$ at positions $\mathcal{K}$, i.e. $y_{\mathcal{K}}=(y_{(\mathcal{K}_{1})},\ldots,~y_{(\mathcal{K}_{|\mathcal{K}|})})^{T}$
; similarly $o_{\mathcal{K}}$ is the vector of values of linear activation
$o$ at positions $\mathcal{K}$, i.e. $o_{\mathcal{K}}=(o_{(\mathcal{K}_{1})},\ldots,~o_{(\mathcal{K}_{|\mathcal{K}|})})^{T}$
. 

Note that the squared error loss belongs to this family as 
\begin{eqnarray*}
\mathcal{\ell}_{\mathrm{squared}} & = & \sum_{j=1}^{D}(o_{j}-y_{j})^{2}\\
 & = & \sum_{j=1}^{D}o_{j}^{2}-2o_{j}y_{j}+y_{j}^{2}\\
 & = & \left(\sum_{j=1}^{D}o_{j}^{2}\right)-2\left(\sum_{j=1}^{D}o_{j}y_{j}\right)+\left(\sum_{j=1}^{D}y_{j}^{2}\right)\\
 & = & \|o\|^{2}-2\left(\sum_{j\in\mathcal{K}}o_{j}y_{j}\right)+\left(\sum_{j\in\mathcal{K}}y_{j}^{2}\right)~\textrm{ since for }j\notin\mathcal{K}\textrm{ we have }y_{j}=0\\
 & = & \|o\|^{2}-2o_{\mathcal{K}}^{T}y_{\mathcal{K}}+\|y_{\mathcal{K}}\|^{2}\\
 & = & \mathcal{\ell}_{\mathrm{squared}}(~\|o\|^{2},~\mathrm{sum}(o),~\mathcal{K},~o_{\mathcal{K}},~y_{\mathcal{K}})
\end{eqnarray*}

where $\mathcal{\ell}_{\mathrm{squared}}$ in particular doesn't use
$\mathrm{sum}(o)$.

The spherical family of loss functions does not include the standard
log of softmax, but it includes possible alternatives, such as the
\emph{spherical softmax} and \emph{Taylor-softmax} that we will introduce
in a later section. Let us detail the steps for computing such a spherical
loss from last hidden layer representation $h$:
\begin{itemize}
\item $o=Wh$
\item $q=\|o\|^{2}=\sum o_{i}^{2}$
\item $s=\mathrm{sum}(o)=\sum o_{i}$
\item $L=\ell(q,s,~\mathcal{K},~o_{\mathcal{K}},~y_{\mathcal{K}})$
\end{itemize}
The gradient of the loss may be backpropagated and the parameters
updated in the \emph{usual naive} way with the following steps: 
\begin{itemize}
\item compute scalars$\frac{\partial\ell}{\partial q}(q,s,~\mathcal{K},~o_{\mathcal{K}},~y_{\mathcal{K}})$
and $\frac{\partial\ell}{\partial s}(q,s,~\mathcal{K},~o_{\mathcal{K}},~y_{\mathcal{K}})$
as well as $K$-dimensional gradient vector $\frac{\partial\ell}{\partial o_{\mathcal{K}}}(q,s,~\mathcal{K},~o_{\mathcal{K}},~y_{\mathcal{K}})$
\item clear $D$-dimensional gradient vector $\nabla_{o}\leftarrow0$
\item update $(\nabla_{o})_{\mathcal{K}}\leftarrow\frac{\partial\ell}{\partial o_{\mathcal{K}}}$
\item update $\nabla_{o}\leftarrow\nabla_{o}+\frac{\partial\ell}{\partial q}\underbrace{\frac{\partial q}{\partial o}}_{2o}$
\item update $\nabla_{o}\leftarrow\nabla_{o}+\frac{\partial\ell}{\partial s}\underbrace{\frac{\partial s}{\partial o}}_{\mathbf{1}_{D}}$
\item backpropagate $\nabla_{h}=W^{T}\nabla_{o}$
\item update $W\leftarrow W-\eta\nabla_{o}h^{T}$ where $\eta$ is a scalar
learning rate.
\end{itemize}
Here again, as in the squared error case, we see that the computation
of $o$ in the forward pass and backpropagation of the gradient to
$\nabla_{h}$ would both require multiplication by the $D\times d$
matrix $W$, and that the update to $W$ will generally be a non-sparse
rank-1 update that requires modifying all its $Dd$ elements. Each
of these three operations have a $O(Dd)$ complexity.

We will now follow the same logical steps as in the simpler squared
error case to derive an efficient algorithm for the spherical loss
family.

\subsection{Efficient computation of the loss}

Let us name the formal parameters of $\ell$ more clearly as follows:

\[
\ell(q,s,\mathcal{K},\mathbf{a},\mathbf{t})
\]

where $q$ and$s$ are scalars that will receive $\|o\|^{2}$ and
$\text{sum}(o)$ respectively; $\mathcal{K}$ is a vector that will
contain the list of at most$K$ indices that correspond to non-zero
elements of sparse $y$; $\mathbf{a}=o_{\mathcal{K}}$ and $\mathbf{t}=y_{\mathcal{K}}$.

\subsubsection{Computing $q=\|o\|^{2}$}

\begin{eqnarray}
q=\|o\|^{2} & = & \|\overbrace{Wh}^{O(Dd)}\|^{2}\nonumber \\
 & = & \left(Wh\right)^{T}\left(Wh\right)\nonumber \\
 & = & h^{T}W^{T}Wh\nonumber \\
 & = & h^{T}(\underbrace{Qh}_{O(d^{2})})\label{eq:q}
\end{eqnarray}

supposing we have maintained an up-to date $Q=W^{T}W$.

Derivative:

\[
\frac{\partial q}{\partial o}=2o
\]

\subsubsection{Computing $s=\mathrm{sum}(o)$}

\begin{eqnarray}
s=\mathrm{sum}(o) & = & \mathrm{sum}(\overbrace{Wh}^{O(Dd)})\nonumber \\
 & = & \sum_{i=1}^{D}\left(\sum_{j=1}^{d}h_{j}W_{j}\right)_{i}\nonumber \\
 & = & \sum_{i=1}^{D}\sum_{j=1}^{d}h_{j}W_{ij}\nonumber \\
 & = & \sum_{j=1}^{d}\left(h_{j}\sum_{i=1}^{D}W_{ij}\right)\nonumber \\
 & = & \sum_{j=1}^{d}h_{j}\underbrace{\mathrm{sum}(W_{j})}_{\bar{w}_{j}}\nonumber \\
 & = & \bar{w}^{T}h\nonumber \\
 & = & h^{T}\bar{w}\label{eq:s}
\end{eqnarray}

This is an $O(d)$ operation, provided we have maintained an up-to-date
vector $\bar{w}=(\mathrm{sum}(W_{1}),~\ldots,~\mathrm{sum}(W_{d}))=W^{T}\mathbf{1}_{D}$.

\[
\frac{\partial s}{\partial o}=\mathbf{1}_{D}
\]

\subsubsection{Computing specific $o_{k}$}

We will also need to compute the specific $o_{k}$ for the few $k\in\mathcal{K}$. 

\begin{eqnarray*}
o_{k} & = & (Wh)_{k}\\
 & = & h^{T}W_{k\bullet}
\end{eqnarray*}

which gives 
\begin{eqnarray}
\mathbf{a}=o_{\mathcal{K}} & = & (o_{(\mathcal{K}_{1})},\ldots,~o_{(\mathcal{K}_{|\mathcal{K}|})})^{T}\nonumber \\
 & = & (h^{T}W_{\mathcal{K}_{1}\bullet},\ldots,~h^{T}W_{\mathcal{K}_{|\mathcal{K}|}\bullet})^{T}\label{eq:avec}
\end{eqnarray}

we then have all we need to pass to loss function $\ell$ to compute
the associated loss

\begin{equation}
L=\ell(q,s,~\mathcal{K},~o_{\mathcal{K}},~y_{\mathcal{K}})=\ell(q,s,~\mathcal{K},~\mathbf{a},~\mathbf{t})\label{eq:onlineL}
\end{equation}

\subsubsection{Corresponding equations for the minibatch case}

In the minibatch case, rather than having the hidden representation
of a single example as a vector $h$ we suppose we receive $m$ hidden
representations in the $m$ columns of a $d\times m$ matrix $H$.
The associated sparse target is $D\times m$ matrix $Y$ whose $m$
columns contain each at most $K$ non-zero elements. $Y$ will be
stored using sparse representation $(\mathcal{K},T)$ where $\mathcal{K}$
is now a $K\times m$ matrix of indices and $T$ is a $K\times m$
matrix containing the corresponding values of $Y$ such that $T_{kj}=Y_{\mathcal{K}_{kj},j}$
for $k\in\{1,\ldots,K\}$ and $j\in\{1,\ldots,m\}$. 

The above equations given for the online case, can easily be adapted
to the minibatch case as follows:

Let $O=WH$ the $D\times m$ matrix of linear outputs whose $j^{th}$
column will contain the output vector of the $j^{th}$ example of
the minibatch. The specific outputs associated to non-zero target
values in $Y$ (whose indexes are in $\mathcal{K}$) will be collected
in $K\times m$ matrix $A$ (the minibatch version of vector $\mathbf{a}$
of Equation \ref{eq:avec} such that 
\begin{equation}
A_{kj}=O_{\mathcal{K}_{kj},j}=(H_{j})^{T}W_{\mathcal{K}_{kj}\bullet}\label{eq:Amat}
\end{equation}

Adapting Equation \ref{eq:q} to the minibatch case, the squared norm
of the $m$ output vectors is obtained in $m$-dimensional vector
$\mathbf{q}$ as

\begin{equation}
\mathbf{q}=\mathrm{diag}(\underbrace{H^{T}\underbrace{QH}_{\hat{H}}}_{\hat{M}})\label{eq:qvec}
\end{equation}

Adapting Equation \ref{eq:s} to the minibatch case, the sum of each
of the $m$ output vectors is obtained in $m$-dimensional vector
$\mathbf{s}$ as

\begin{equation}
\mathbf{s}=H^{T}\bar{w}\label{eq:svec}
\end{equation}

Adapting Equation \ref{eq:onlineL} the corresponding vector of $m$
individual losses for the $m$ examples of the minibatch is 
\begin{equation}
\vec{L}=\left[\ell(\mathbf{q}_{j},\mathbf{s}_{j},\mathcal{K}_{j,},A_{j},T_{j})\right]_{j=1\ldots m}\label{eq:Lvec}
\end{equation}

and the total loss for the minibatch is 
\begin{equation}
L=\mathrm{sum}(\vec{L})\label{eq:gen-L-minibatch-sum}
\end{equation}

\subsection{Gradient of loss $L$ with respect to $h$}

\subsubsection*{Online case:}

To backpropagate the gradients through the network, we first need
the gradients with respect to linear activations $o$: $\nabla_{o}=\frac{\partial L}{\partial o}$.

There will be three types of contributions to this gradient: contribution
due to $q$, contribution due to $s$, and contribution due to \emph{direct}
influence on the loss of the $o_{k}$ for $k\in\mathcal{K}$. 

\begin{eqnarray*}
\nabla_{o}=\frac{\partial L}{\partial o} & = & \frac{\partial\ell}{\partial q}\frac{\partial q}{\partial o}+\frac{\partial\ell}{\partial s}\frac{\partial s}{\partial o}+\sum_{k=1}^{K}\frac{\partial\ell}{\partial\mathbf{a}_{k}}\frac{\partial\mathbf{a}_{k}}{\partial o}
\end{eqnarray*}

We have $\frac{\partial q}{\partial o}=2o$, $\frac{\partial s}{\partial o}=\mathbf{1}_{D}$
and $\frac{\partial\mathbf{a}_{k}}{\partial o}=\mathrm{onehot}_{D}(\mathcal{K}_{k})$
because $\mathbf{a}_{k}=o_{\mathcal{K}_{k}}$ so this becomes

\begin{eqnarray}
\nabla_{o} & = & 2o~\frac{\partial\ell}{\partial q}+\mathbf{1}_{D}\frac{\partial\ell}{\partial s}+\sum_{k=1}^{K}\frac{\partial\ell}{\partial\mathbf{a}_{k}}\mathrm{onehot}_{D}(\mathcal{K}_{k})\nonumber \\
 & = & 2o~\frac{\partial\ell}{\partial q}+\mathbf{1}_{D}\frac{\partial\ell}{\partial s}+\mathring{y}\label{eq:grad_o_online}
\end{eqnarray}

where we have defined vector $\mathring{y}=\sum_{k=1}^{K}\frac{\partial\ell}{\partial\mathbf{a}_{k}}\mathrm{onehot}_{D}(\mathcal{K}_{k})$
as a sparse vector, having value at position $\mathbf{k}_{j}$ equal
$\frac{\partial\ell}{\partial\mathbf{a}_{j}}$. It will, like $y$,
be stored in $K-sparse$ representation, with the indexes given by
$\mathbf{k}$ and the corresponding values in $\frac{\partial\ell}{\partial\mathbf{a}_{j}}$.

Gradient with respect to $h$:

\begin{eqnarray*}
\nabla_{h} & = & \frac{\partial o}{\partial h}\frac{\partial L}{\partial o}\\
 & = & W^{T}\nabla_{o}\\
 & = & W^{T}\left(2o~\frac{\partial\ell}{\partial q}+\mathbf{1}_{D}\frac{\partial\ell}{\partial s}+\mathring{y}\right)\\
 & = & 2W^{T}o~\frac{\partial\ell}{\partial q}+W^{T}\mathbf{1}_{D}\frac{\partial\ell}{\partial s}+W^{T}\mathring{y}\\
 & = & 2W^{T}Wh~\frac{\partial\ell}{\partial q}+\bar{w}\frac{\partial\ell}{\partial s}+W^{T}\mathring{y}\\
 & = & 2Qh~\frac{\partial\ell}{\partial q}+\bar{w}\frac{\partial\ell}{\partial s}+W^{T}\sum_{k=1}^{K}\frac{\partial\ell}{\partial\mathbf{a}_{k}}\mathrm{onehot}_{D}(\mathcal{K}_{k})\\
 & = & 2Qh~\frac{\partial\ell}{\partial q}+\bar{w}\frac{\partial\ell}{\partial s}+\sum_{k=1}^{K}\frac{\partial\ell}{\partial\mathbf{a}_{k}}W^{T}\mathrm{onehot}_{D}(\mathcal{K}_{k})\\
 & = & 2Qh~\frac{\partial\ell}{\partial q}+\bar{w}\frac{\partial\ell}{\partial s}+\sum_{k=1}^{K}\frac{\partial\ell}{\partial\mathbf{a}_{k}}W_{\mathcal{K}_{k}\bullet}
\end{eqnarray*}

\subsubsection*{Minibatch case:}

We now consider a minibatch of $m$ examples whose corresponding linear
outputs are in a $D\times m$ matrix $O=WH$. Let us also denote the
vectors of gradients of the loss with respect to $\mathbf{q}$ and
$\mathbf{s}$ as:

\begin{eqnarray*}
\nabla_{q} & = & \left[\frac{\partial\ell}{\partial q}(\mathbf{q}_{j},\mathbf{s}_{j},\mathcal{K}_{j,},A_{j},T_{j})\right]_{j=1\ldots m}\\
\nabla_{s} & = & \left[\frac{\partial\ell}{\partial s}(\mathbf{q}_{j},\mathbf{s}_{j},\mathcal{K}_{j,},A_{j},T_{j})\right]_{j=1\ldots m}
\end{eqnarray*}

Let us also define

\[
\nabla_{A}=\left[\frac{\partial\ell}{\partial\mathbf{a}_{k}}(\mathbf{q}_{j},\mathbf{s}_{j},\mathcal{K}_{j,},A_{j},T_{j})\right]_{k=1\ldots K,~j=1\ldots m}
\]

and $\mathring{Y}$ as the \uline{sparse} $D\times m$ whose column
$j$ is defined as 
\begin{eqnarray*}
\mathring{Y}_{j} & = & \sum_{k=1}^{K}\frac{\partial\ell}{\partial\mathbf{a}_{k}}(\mathbf{q}_{j},\mathbf{s}_{j},\mathcal{K}_{j,},A_{j},T_{j})~\mathrm{onehot}_{D}(\mathcal{K}_{kj})\\
 & = & \sum_{k=1}^{K}(\nabla_{A})_{kj}~\mathrm{onehot}_{D}(\mathcal{K}_{kj})
\end{eqnarray*}
 which may be summarize as$\mathring{Y}_{\mathcal{K}_{j}}=\left(\nabla_{A}\right)_{j}$ 

Equation \ref{eq:grad_o_online} then becomes in the minibatch case:

\begin{eqnarray*}
\nabla_{O_{j}} & = & 2O_{j}~\left(\nabla_{q}\right)_{j}+\mathbf{1}_{D}\left(\nabla_{s}\right)_{j}+\mathring{Y}_{j}
\end{eqnarray*}

or in matrix form

\begin{eqnarray}
\nabla_{O} & = & 2O~\mathrm{diag}(\nabla_{q})+\mathbf{1}_{D}\nabla_{s}^{T}+\mathring{Y}\label{eq:gen-grad-O}
\end{eqnarray}

and the gradient with respect to $H$ is:

\begin{eqnarray}
\nabla_{H} & = & \frac{\partial L}{\partial H}\nonumber \\
 & = & \frac{\partial O}{\partial H}\frac{\partial L}{\partial O}\nonumber \\
 & = & W^{T}\nabla_{O}\label{eq:gen-gradH-highlevel}\\
 & = & W^{T}\left(2O~\mathrm{diag}(\nabla_{q})+\mathbf{1}_{D}\nabla_{s}^{T}+\mathring{Y}\right)\nonumber \\
 & = & 2W^{T}O~\mathrm{diag}(\nabla_{q})+W^{T}\mathbf{1}_{D}\nabla_{s}^{T}+W^{T}\mathring{Y}\nonumber \\
 & = & 2W^{T}WH~\mathrm{diag}(\nabla_{q})+\bar{w}\nabla_{s}^{T}+W^{T}\mathring{Y}\nonumber \\
 & = & 2QH~\mathrm{diag}(\nabla_{q})+\underbrace{\bar{w}\nabla_{s}^{T}+W^{T}\mathring{Y}}_{\hat{Z}}\label{eq:gen-gradH-unfact}
\end{eqnarray}

where we define the $d\times m$ matrix $\hat{Z}$ as 
\begin{equation}
\hat{Z}=\bar{w}\nabla_{s}^{T}+W^{T}\mathring{Y}\label{eq:gen_Zhat_nofact}
\end{equation}

\subsection{Standard naive gradient update of parameters $W$ }

The gradient of the loss with respect to output layer weight matrix
$W$ is 

\begin{eqnarray*}
\frac{\partial L}{\partial W} & = & \frac{\partial L}{\partial O}\frac{\partial O}{\partial W}\\
 & = & \nabla_{O}H^{T}\\
 & = & \left(2O~\mathrm{diag}(\nabla_{q})+\mathbf{1}_{D}\nabla_{s}^{T}+\mathring{Y}\right)H^{T}\\
 & = & \left(2WH~\mathrm{diag}(\nabla_{q})+\mathbf{1}_{D}\nabla_{s}^{T}+\mathring{Y}\right)H^{T}
\end{eqnarray*}

And the corresponding gradient descent update to $W$ would thus be
\begin{eqnarray}
W_{new} & = & W-\eta\left(2WH~\mathrm{diag}(\nabla_{q})+\mathbf{1}_{D}\nabla_{s}^{T}+\mathring{Y}\right)H^{T}\label{eq:gen-update-W}
\end{eqnarray}

where $\eta$ is a positive learning rate. 

Computed in this manner, this induces a prohibitive $O(mDd)$ computational
complexity, first to compute $WH$, and then to update all the $Dd$
elements of $W$. Note that all $D\times d$ elements of $W$ must
be accessed during this update. On the surface this seems hopeless.
But we will see in the next section how we can achieve the \emph{exact}
same update of $W$ in $O(md^{2})$.

\subsection{Efficient gradient update of parameters using a factored representation
of $W$\label{sub:gen-efficient-gradient-update}}

First note that the update of $W$ given in equation \ref{eq:gen-update-W}
can be decomposed in 3 consecutive updates:

\begin{eqnarray*}
\mathbf{a)}\,\,W & \leftarrow & W-2\eta(WH)~\mathrm{diag}(\nabla_{q})H^{T}\\
\mathbf{b)}\,\,W & \leftarrow & W-\eta\mathbf{1}_{D}\nabla_{s}^{T}H^{T}\\
\mathbf{c)}\,\,W & \leftarrow & W-\eta\mathring{Y}H^{T}
\end{eqnarray*}

In doing this we haven't yet changed anything to the $O(mDd)$ complexity
of this update. Note that update a) can also be seen as $W\leftarrow W\left(\mathbf{I}-2\eta H~\mathrm{diag}(\nabla_{q})H^{T}\right)$.

The trick now is to represent $W$ implicitly as\footnote{Note that we never actually \emph{factorize} an arbitrary pre-exisitng
$W$, which would be prohibitive as $W$ is huge. We will no longer
store or update a $W$, but only$V,U,\omega$ which implicitly represent
$W$.}: 
\begin{equation}
\underbrace{W}_{D\times d}=\underbrace{V}_{D\times d}\underbrace{U}_{d\times d}+\mathbf{1}_{D}\omega^{T}\label{eq:gen-Wfact}
\end{equation}
 where $\omega$ is a $d$-dimensional vector. In this case the following
updates to $V,U,\omega$ respectively will implicitly update the implicit
$W$ in the exact same way as the above 3 updates:

\begin{eqnarray}
\mathbf{a)}\,\,U_{new} & = & U\left(\mathbf{I}-2\eta H~\mathrm{diag}(\nabla_{q})H^{T}\right)\nonumber \\
 & = & U-2\eta UH~\mathrm{diag}(\nabla_{q})H^{T}\label{eq:gen-updateU}\\
\mathbf{b)}\,\,\omega_{new} & = & \left(\mathbf{I}-2\eta H~\mathrm{diag}(\nabla_{q})H^{T}\right)^{T}\omega-\eta H\nabla_{s}\nonumber \\
 & = & \omega-2\eta H~\mathrm{diag}(\nabla_{q})H^{T}\omega-\eta H\nabla_{s}\nonumber \\
 & = & \omega-\eta H\left(2~\mathrm{diag}(\nabla_{q})H^{T}\omega+\nabla_{s}\right)\label{eq:gen-update-omega}\\
\mathbf{c)}\,\,V_{new} & = & V-\eta\mathring{Y}(U_{new}^{-T}H){}^{T}\label{eq:gen-updateV}
\end{eqnarray}

But, with this formulation, provided we keep an up-to-date $U^{-T}$
(which we will see we can do cheaply using the Woodbury identity),
the whole update to $V,U,\omega$ is now $O(md^{2})$ rather than
the equivalent naive $O(mDd)$ update of Eq. \ref{eq:gen-update-W}
to an explicit $W$.

Indeed, step a) and b) involve only multiplications between matrices
of dimensions $d\times m$ and $d\times d$ (matrices $H$ and $U$).
As for step c) it involves an $O(md^{2})$ multiplication of $U^{-T}$
by $H$, followed by a \textbf{sparse update} of $V$. Since $\mathring{Y}$
is an extremely sparse $D\times m$ matrix whose $m$ columns each
contain at most $K$ non-zero elements, update c) will touch at most
$Km$ rows of $V$, yielding an $O(Kmd)$ operation. This is to be
contrasted with the standard, equivalent but naive update of Eq. \ref{eq:gen-update-W}
to an explicit $W$, which requires accessing and modifying all $D\times d$
elements of $W$ for every update and yields an overall $O(mDd)$
computational complexity.

\subsubsection*{Proof that this sequence of updates yields the update of W given
above:}

\begin{eqnarray*}
 &  & V_{new}U_{new}+\mathbf{1}_{D}\omega_{new}^{T}\\
 & = & \left(V-\eta\mathring{Y}(U_{new}^{-T}H){}^{T}\right)U_{new}+\mathbf{1}_{D}\left(\omega-\eta H\left(2~\mathrm{diag}(\nabla_{q})H^{T}\omega+\nabla_{s}\right)\right)^{T}\\
 & = & \left(V-\eta\mathring{Y}H^{T}U_{new}^{-1}\right)U_{new}+\mathbf{1}_{D}\left(\omega^{T}-\eta\left(2~\mathrm{diag}(\nabla_{q})H^{T}\omega+\nabla_{s}\right)^{T}H^{T}\right)\\
 & = & VU_{new}-\eta\mathring{Y}H^{T}U_{new}^{-1}U_{new}+\mathbf{1}_{D}\omega^{T}-\eta\mathbf{1}_{D}\left(2~\mathrm{diag}(\nabla_{q})H^{T}\omega+\nabla_{s}\right)^{T}H^{T}\\
 & = & VU_{new}-\eta\mathring{Y}H^{T}+\mathbf{1}_{D}\omega^{T}-\eta\mathbf{1}_{D}\left(2~\mathrm{diag}(\nabla_{q})H^{T}\omega+\nabla_{s}\right)^{T}H^{T}\\
 & = & V\left(U-2\eta UH~\mathrm{diag}(\nabla_{q})H^{T}\right)-\eta\mathring{Y}H^{T}+\mathbf{1}_{D}\omega^{T}-\eta\mathbf{1}_{D}\left(2~\mathrm{diag}(\nabla_{q})H^{T}\omega+\nabla_{s}\right)^{T}H^{T}\\
 & = & VU-2\eta VUH~\mathrm{diag}(\nabla_{q})H^{T}-\eta\mathring{Y}H^{T}+\mathbf{1}_{D}\omega^{T}-\eta\mathbf{1}_{D}\left(2~\omega^{T}H~\mathrm{diag}(\nabla_{q})+\nabla_{s}^{T}\right)H^{T}\\
 & = & (VU+\mathbf{1}_{D}\omega^{T})-2\eta VUH~\mathrm{diag}(\nabla_{q})H^{T}-\eta\mathring{Y}H^{T}-\eta\mathbf{1}_{D}\left(2~\omega^{T}H\mathrm{~diag}(\nabla_{q})+\nabla_{s}^{T}\right)H^{T}\\
 & = & W-2\eta VUH~\mathrm{diag}(\nabla_{q})H^{T}-\eta\mathring{Y}H^{T}-2\eta\mathbf{1}_{D}\omega^{T}H~\mathrm{diag}(\nabla_{q})H^{T}-\eta\mathbf{1}_{D}\nabla_{s}^{T}H^{T}\\
 & = & W-2\eta VUH~\mathrm{diag}(\nabla_{q})H^{T}-2\eta\mathbf{1}_{D}\omega^{T}H~\mathrm{diag}(\nabla_{q})H^{T}-\eta\mathbf{1}_{D}\nabla_{s}^{T}H^{T}-\eta\mathring{Y}H^{T}\\
 & = & W-2\eta\left(VUH~\mathrm{diag}(\nabla_{q})H^{T}+\mathbf{1}_{D}\omega^{T}H~\mathrm{diag}(\nabla_{q})H^{T}\right)-\eta\mathbf{1}_{D}\nabla_{s}^{T}H^{T}-\eta\mathring{Y}H^{T}\\
 & = & W-2\eta\left(VU+\mathbf{1}_{D}\omega^{T}\right)H~\mathrm{diag}(\nabla_{q})H^{T}-\eta\mathbf{1}_{D}\nabla_{s}^{T}H^{T}-\eta\mathring{Y}H^{T}\\
 & = & W-2\eta WH~\mathrm{diag}(\nabla_{q})H^{T}-\eta\mathbf{1}_{D}\nabla_{s}^{T}H^{T}-\eta\mathring{Y}H^{T}\\
 & = & W-\eta\left(2WH~\mathrm{diag}(\nabla_{q})+\mathbf{1}_{D}\nabla_{s}^{T}+\mathring{Y}\right)H^{T}\\
 & = & W_{new}
\end{eqnarray*}

\subsection{Adapting the computation of loss $L$ and gradient $\nabla_{H}$
to the factorized representation\label{sub:gen-adapting-L-graH-factored}}

Let us now adapt the computation of loss $L$ and gradient $\nabla_{H}$
now that we no longer have an explicit $W$ but rather store it implicitly
as $W=VU+\mathbf{1}_{D}\omega^{T}$.

\subsubsection{Loss $L$}

Computing the total loss $L$ over a minibatch implies computing $L=\mathrm{sum}(\vec{L})=\mathrm{sum}\left(\left[\ell(\mathbf{q}_{j},\mathbf{s}_{j},\mathcal{K}_{j,},A_{j},T_{j})\right]_{j=1\ldots m}\right)$
as previously seen in Eq. \ref{eq:Lvec} and Eq. \ref{eq:gen-L-minibatch-sum}.
Index matrix $\mathcal{K}$ and associated target matrix $T$ are
the same as before. Vectors $\mathbf{q}$ and $\mathbf{s}$ can be
computed cheaply as previously using Eq. \ref{eq:qvec} and \ref{eq:svec}
provided we have kept an up-to-date $Q$ and $\bar{w}$ (we shall
see how to update them effectively in the next section). So to be
able to compute loss $L$ using this factored representation of $W$
it remains only to adapt the computation of $K\times m$ matrix $A$.
This matrix was defined in Eq. \ref{eq:Amat} as $A_{kj}=O_{\mathcal{K}_{kj},j}=(H_{j})^{T}W_{\mathcal{K}_{kj}\bullet}$.
Replacing $W$ by its factored expression we can write

\begin{eqnarray*}
A_{kj} & = & (H_{j})^{T}\left(VU+\mathbf{1}_{D}\omega^{T}\right)_{\mathcal{K}_{kj}\bullet}\\
 & = & (H_{j})^{T}\left(VU\right)_{\mathcal{K}_{kj}\bullet}+(H_{j})^{T}\left(\mathbf{1}_{D}\omega^{T}\right)_{\mathcal{K}_{kj}\bullet}\\
 & = & (H_{j})^{T}\left(VU\right)_{\mathcal{K}_{kj}\bullet}+(H_{j})^{T}\omega\\
 & = & (H_{j})^{T}\left((VU)^{T}\right)_{\mathcal{K}_{kj}}+(H_{j})^{T}\omega\\
 & = & (H_{j})^{T}\left(U^{T}V^{T}\right)_{\mathcal{K}_{kj}}+(H_{j})^{T}\omega\\
 & = & (H_{j})^{T}U^{T}\left(V^{T}\right)_{\mathcal{K}_{kj}}+(H_{j})^{T}\omega\\
 & = & (UH_{j})^{T}\left(V^{T}\right)_{\mathcal{K}_{kj}}+(H_{j})^{T}\omega\\
 & = & ((UH)_{j})^{T}V_{\mathcal{K}_{kj}\bullet}+(H_{j})^{T}\omega\\
 & = & ((\underbrace{UH}_{\tilde{H}})_{j})^{T}V_{\mathcal{K}_{kj}\bullet}+(\underbrace{H{}^{T}\omega}_{\tilde{\mathbf{h}}})_{j}
\end{eqnarray*}

In summary, having computed

\begin{equation}
\tilde{H}=UH\label{eq:gen-Htilde}
\end{equation}
 and

\begin{equation}
\tilde{\mathbf{h}}=H^{T}\omega\label{eq:gen-htilde}
\end{equation}
we can efficiently compute the elements of $K\times m$ matrix $A$
by accessing only the rows of $V$ whose indexes are in $\mathcal{K}$as
follows: 
\begin{equation}
A_{kj}=(\tilde{H}_{j})^{T}V_{\mathcal{K}_{kj}\bullet}+\tilde{\mathbf{h}}_{j}\label{eq:gen-A-fact}
\end{equation}

\subsubsection{Gradient $\nabla_{H}$}

\label{sub:gen-grad-H-fact}

Let us now adapt the computation of the gradient with respect to $H$,
starting from previous Eq. \ref{eq:gen-gradH-unfact} i.e.$\nabla_{H}=2QH~\mathrm{diag}(\nabla_{q})+\hat{Z}$
with $\hat{Z}=\bar{w}\nabla_{s}^{T}+W^{T}\mathring{Y}$. 

Supposing we have kept an up-to-date $Q$ and $\bar{w}$ (we shall
see how to update them effectively in the section \ref{sub:gen-bookkeeping-updating-wbar-and-Q}),
we are left with only adapting the computation of the $W^{T}\mathring{Y}$
term to use the factored representation of $W$:

\begin{eqnarray}
\hat{Z} & = & \bar{w}\nabla_{s}^{T}+W^{T}\mathring{Y}\nonumber \\
 & = & \bar{w}\nabla_{s}^{T}+\left(VU+\mathbf{1}_{D}\omega^{T}\right)^{T}\mathring{Y}\nonumber \\
 & = & \bar{w}\nabla_{s}^{T}+U^{T}V^{T}\mathring{Y}+\omega\mathbf{1}_{D}^{T}\mathring{Y}\nonumber \\
 & = & \bar{w}\nabla_{s}^{T}+U^{T}(V^{T}\mathring{Y})+\omega(\mathring{Y}^{T}\mathbf{1}_{D})^{T}\nonumber \\
 & = & \bar{w}\nabla_{s}^{T}+U^{T}(V^{T}\mathring{Y})+\omega\mathbf{\bar{y}}^{T}\label{eq:gen-Zhat-fact}
\end{eqnarray}

provided we defined 
\begin{equation}
\mathbf{\bar{y}}=\mathring{Y}^{T}\mathbf{1}_{D}=\mathrm{rowsum}(\mathring{Y})=\mathrm{rowsum}(\nabla_{A})\label{eq:ybar}
\end{equation}

We see that computing $d\times m$ matrix $\hat{Z}$ in this manner
can be achieved efficiently using our factored representation $V,U$
and $\omega$. Note that computing $V^{T}\mathring{Y}$ is a multiplication
by sparse matrix $\mathring{Y}$ which will have a computational complexity
of $O(Kdm)$, and yield a $d\times m$ matrix. The computation of
$\hat{Z}$ in this manner thus has a$O(dm+d^{2}m+Kdm+dm)$ complexity. 

We can then proceed to computing $\nabla_{H}$ as in Eq. \ref{eq:gen-gradH-unfact}:
\begin{equation}
\nabla_{H}=2\underbrace{QH}_{\hat{H}}~\mathrm{diag}(\nabla_{q})+\hat{Z}\label{eq:gen-grad-H-fact}
\end{equation}

\subsection{Bookkeeping operations: keeping up-to-date $\bar{w}$ and $Q$}

\label{sub:gen-bookkeeping-updating-wbar-and-Q}

We have shown in section \ref{sub:gen-efficient-gradient-update}
that our updates to $V,U,\omega$ (Eq. \ref{eq:gen-updateV},\ref{eq:gen-updateU},\ref{eq:gen-update-omega})
achieve the same update on (an implicit) $W$ as Eq. \ref{eq:gen-update-W},
i.e. $W_{new}=W-\eta\left(2WH~\mathrm{diag}(\nabla_{q})+\mathbf{1}_{D}\nabla_{s}^{T}+\mathring{Y}\right)H^{T}$.
The efficient computation of loss $L$ and gradient $\nabla_{H}$
seen in Section \ref{sub:gen-adapting-L-graH-factored} relies on
having an up-to-date $Q=W^{T}W$ and $\bar{w}=\mathrm{rowsum}(W)=(\mathrm{sum}(W_{1}),~\ldots,~\mathrm{sum}(W_{d}))=W^{T}\mathbf{1}_{D}$.
In this section, we derive efficient updates to $\bar{w}$ and $Q$
that reflect the update to $W$.

\subsubsection{Update of $\bar{w}$ }

\begin{eqnarray}
\bar{w}_{new} & = & W_{new}^{T}\mathbf{1}_{D}\nonumber \\
 & = & \left(W-\eta\left(2WH~\mathrm{diag}(\nabla_{q})+\mathbf{1}_{D}\nabla_{s}^{T}+\mathring{Y}\right)H^{T}\right)^{T}\mathbf{1}_{D}\nonumber \\
 & = & W^{T}\mathbf{1}_{D}-\eta H\left(2WH~\mathrm{diag}(\nabla_{q})+\mathbf{1}_{D}\nabla_{s}^{T}+\mathring{Y}\right)^{T}\mathbf{1}_{D}\nonumber \\
 & = & \bar{w}-\eta H\left(2\mathrm{diag}(\nabla_{q})H^{T}W^{T}+\nabla_{s}\mathbf{1}_{D}^{T}+\mathring{Y}^{T}\right)\mathbf{1}_{D}\nonumber \\
 & = & \bar{w}-2\eta H\mathrm{diag}(\nabla_{q})H^{T}W^{T}\mathbf{1}_{D}-\eta H\nabla_{s}\mathbf{1}_{D}^{T}\mathbf{1}_{D}-\eta H\underbrace{\mathring{Y}^{T}\mathbf{1}_{D}}_{\bar{y}}\nonumber \\
 & = & \bar{w}-2\eta H\mathrm{diag}(\nabla_{q})H^{T}\bar{w}-\eta DH\nabla_{s}-\eta H\mathbf{\bar{y}}\nonumber \\
 & = & \bar{w}-\eta H\left(2\mathrm{diag}(\nabla_{q})H^{T}\bar{w}-\eta D\nabla_{s}-\eta\mathbf{\bar{y}}\right)\label{eq:gen-update-wbar}
\end{eqnarray}

\subsubsection{Update of $Q$ }

\begin{eqnarray}
Q_{new} & = & W_{new}^{T}W_{new}\nonumber \\
 & = & \left(W-\eta\nabla_{O}H^{T}\right)^{T}\left(W-\eta\nabla_{O}H^{T}\right)\nonumber \\
 & = & W^{T}W-W^{T}\left(\eta\nabla_{O}H^{T}\right)-\left(\eta\nabla_{O}H^{T}\right)^{T}W+\eta^{2}\left(\nabla_{O}H^{T}\right)^{T}\nabla_{O}H^{T}\nonumber \\
 & = & \underbrace{W^{T}W}_{Q}-\eta\underbrace{W^{T}\nabla_{O}}_{\nabla_{H}}H^{T}-\eta(\underbrace{W^{T}\nabla_{O}}_{\nabla_{H}}H^{T})^{T}+\eta^{2}H\nabla_{O}^{T}\nabla_{O}H^{T}\nonumber \\
Q_{new} & = & Q-\eta\left(\nabla_{H}H^{T}\right)-\eta\left(\nabla_{H}H^{T}\right)^{T}+\eta^{2}H(\underbrace{\nabla_{O}^{T}\nabla_{O}}_{M})H^{T}\label{eq:gen-Qnew}
\end{eqnarray}

where we used the fact that $W^{T}W=Q$ and $\nabla_{H}=W^{T}\nabla_{O}$.
Note that while computing $\nabla_{O}H^{T}$ would be a prohibitive
$O(mDd)$ computation (in addition to requiring to explicitly compute
$\nabla_{O}$ in the first place), computing $\nabla_{H}H^{T}$ is
a comparatively cheap $O(md^{2})$ operation.

It remains to derive a way to efficiently compute $m\times m$ matrix
$M=\nabla_{O}^{T}\nabla_{O}$ without explicitly computing $O$ nor
resorting to explicit $W$. Substituting $\nabla_{O}$ by its expression
from Eq. \ref{eq:gen-grad-O} i.e. $\nabla_{O}=2O~\mathrm{diag}(\nabla_{q})+\mathbf{1}_{D}\nabla_{s}^{T}+\mathring{Y}$
yields

\begin{eqnarray*}
M & = & \nabla_{O}^{T}\nabla_{O}\\
M & = & \left(2O~\mathrm{diag}(\nabla_{q})+\mathbf{1}_{D}\nabla_{s}^{T}+\mathring{Y}\right)^{T}\left(2O~\mathrm{diag}(\nabla_{q})+\mathbf{1}_{D}\nabla_{s}^{T}+\mathring{Y}\right)\\
M & = & \left(\left(2O~\mathrm{diag}(\nabla_{q})\right)^{T}+\left(\mathbf{1}_{D}\nabla_{s}^{T}+\mathring{Y}\right)^{T}\right)\left(\left(2O~\mathrm{diag}(\nabla_{q})\right)+\left(\mathbf{1}_{D}\nabla_{s}^{T}+\mathring{Y}\right)\right)\\
M & = & \left(2O~\mathrm{diag}(\nabla_{q})\right)^{T}\left(2O~\mathrm{diag}(\nabla_{q})\right)+\left(\mathbf{1}_{D}\nabla_{s}^{T}+\mathring{Y}\right)^{T}\left(\mathbf{1}_{D}\nabla_{s}^{T}+\mathring{Y}\right)\\
 &  & +\left(2O~\mathrm{diag}(\nabla_{q})\right)^{T}\left(\mathbf{1}_{D}\nabla_{s}^{T}+\mathring{Y}\right)+\left(\mathbf{1}_{D}\nabla_{s}^{T}+\mathring{Y}\right)^{T}\left(2O~\mathrm{diag}(\nabla_{q})\right)\\
M & = & \left(4\mathrm{diag}(\nabla_{q})O^{T}O~\mathrm{diag}(\nabla_{q})\right)+\left(\nabla_{s}\mathbf{1}_{D}^{T}\mathbf{1}_{D}\nabla_{s}^{T}+\mathring{Y}^{T}\mathring{Y}+\nabla_{s}\mathbf{1}_{D}^{T}\mathring{Y}+\mathring{Y}^{T}\mathbf{1}_{D}\nabla_{s}^{T}\right)\\
 &  & +\left(2O~\mathrm{diag}(\nabla_{q})\right)^{T}\left(\mathbf{1}_{D}\nabla_{s}^{T}+\mathring{Y}\right)+\left(\mathbf{1}_{D}\nabla_{s}^{T}+\mathring{Y}\right)^{T}\left(2O~\mathrm{diag}(\nabla_{q})\right)\\
M & = & 4\mathrm{diag}(\nabla_{q})O^{T}O~\mathrm{diag}(\nabla_{q})+\left(D\nabla_{s}\nabla_{s}^{T}+\mathring{Y}^{T}\mathring{Y}+\nabla_{s}\mathbf{\bar{y}}^{T}+\mathbf{\bar{y}}\nabla_{s}^{T}\right)\\
 &  & +\left(2O~\mathrm{diag}(\nabla_{q})\right)^{T}\left(\nabla_{s}\mathbf{1}_{D}^{T}+\mathring{Y}^{T}\right)^{T}+\left(\nabla_{s}\mathbf{1}_{D}^{T}+\mathring{Y}^{T}\right)\left(2O~\mathrm{diag}(\nabla_{q})\right)\\
M & = & 4\mathrm{diag}(\nabla_{q})O^{T}O~\mathrm{diag}(\nabla_{q})+\left(D\nabla_{s}\nabla_{s}^{T}+\mathring{Y}^{T}\mathring{Y}+\nabla_{s}\mathbf{\bar{y}}^{T}+\mathbf{\bar{y}}\nabla_{s}^{T}\right)\\
 &  & +\left(\left(\nabla_{s}\mathbf{1}_{D}^{T}+\mathring{Y}^{T}\right)\left(2O~\mathrm{diag}(\nabla_{q})\right)\right)^{T}+\left(\left(\nabla_{s}\mathbf{1}_{D}^{T}+\mathring{Y}^{T}\right)\left(2O~\mathrm{diag}(\nabla_{q})\right)\right)\\
M & = & 4\mathrm{diag}(\nabla_{q})O^{T}O~\mathrm{diag}(\nabla_{q})+\left(D\nabla_{s}\nabla_{s}^{T}+\mathring{Y}^{T}\mathring{Y}+\nabla_{s}\mathbf{\bar{y}}^{T}+\mathbf{\bar{y}}\nabla_{s}^{T}\right)\\
 &  & +\left(2\nabla_{s}\mathbf{1}_{D}^{T}O~\mathrm{diag}(\nabla_{q})+2\mathring{Y}^{T}O~\mathrm{diag}(\nabla_{q})\right)^{T}+\left(2\nabla_{s}\mathbf{1}_{D}^{T}O~\mathrm{diag}(\nabla_{q})+2\mathring{Y}^{T}O~\mathrm{diag}(\nabla_{q})\right)
\end{eqnarray*}

Since $O=WH$ we have $O^{T}O=H^{T}W^{T}WH=H^{T}QH$ and $\mathbf{1}_{D}^{T}O=\mathbf{1}_{D}^{T}WH=\bar{w}^{T}H$.
Substituting these in the above expression of $M$ we obtain

\begin{eqnarray*}
M & = & 4\mathrm{diag}(\nabla_{q})\overbrace{O^{T}O}^{H^{T}QH}~\mathrm{diag}(\nabla_{q})+(D\nabla_{s}\nabla_{s}^{T}+\overbrace{\mathring{Y}^{T}\mathring{Y}}^{\mathring{M}}+\nabla_{s}\mathbf{\bar{y}}^{T}+\mathbf{\bar{y}}\nabla_{s}^{T})\\
 &  & +(2\nabla_{s}\underbrace{\mathbf{1}_{D}^{T}O}_{\bar{w}^{T}H}~\mathrm{diag}(\nabla_{q})+2\mathring{Y}^{T}\underbrace{O}_{WH}~\mathrm{diag}(\nabla_{q}))^{T}+(2\nabla_{s}\underbrace{\mathbf{1}_{D}^{T}O}_{\bar{w}^{T}H}~\mathrm{diag}(\nabla_{q})+2\mathring{Y}^{T}\underbrace{O}_{WH}~\mathrm{diag}(\nabla_{q}))\\
\\
M & = & 4\mathrm{diag}(\nabla_{q})H^{T}QH~\mathrm{diag}(\nabla_{q})+\left(D\nabla_{s}\nabla_{s}^{T}+\mathring{Y}^{T}\mathring{Y}+\nabla_{s}\mathbf{\bar{y}}^{T}+\mathbf{\bar{y}}\nabla_{s}^{T}\right)\\
 &  & +2\left(\nabla_{s}\bar{w}^{T}H~\mathrm{diag}(\nabla_{q})+\mathring{Y}^{T}WH~\mathrm{diag}(\nabla_{q})\right)^{T}+2\left(\nabla_{s}\bar{w}^{T}H~\mathrm{diag}(\nabla_{q})+\mathring{Y}^{T}WH~\mathrm{diag}(\nabla_{q})\right)\\
M & = & 4\mathrm{diag}(\nabla_{q})H^{T}QH~\mathrm{diag}(\nabla_{q})+\left(D\nabla_{s}\nabla_{s}^{T}+\mathring{Y}^{T}\mathring{Y}+\nabla_{s}\mathbf{\bar{y}}^{T}+\mathbf{\bar{y}}\nabla_{s}^{T}\right)\\
 &  & +2\left(\left(\nabla_{s}\bar{w}^{T}+\mathring{Y}^{T}W\right)H~\mathrm{diag}(\nabla_{q})\right)^{T}+2\left(\left(\nabla_{s}\bar{w}^{T}+\mathring{Y}^{T}W\right)H~\mathrm{diag}(\nabla_{q})\right)\\
M & = & 4\mathrm{diag}(\nabla_{q})H^{T}QH~\mathrm{diag}(\nabla_{q})+\left(D\nabla_{s}\nabla_{s}^{T}+\mathring{Y}^{T}\mathring{Y}+\nabla_{s}\mathbf{\bar{y}}^{T}+\mathbf{\bar{y}}\nabla_{s}^{T}\right)\\
 &  & +2(\mathrm{diag}(\nabla_{q})H^{T}\underbrace{(\bar{w}\nabla_{s}^{T}+W^{T}\mathring{Y})}_{\hat{Z}}+2(\mathrm{diag}(\nabla_{q})H^{T}\underbrace{(\bar{w}\nabla_{s}^{T}+W^{T}\mathring{Y})}_{\hat{Z}})^{T}.
\end{eqnarray*}

Reusing previously defined $\hat{Z}=\bar{w}\nabla_{s}^{T}+W^{T}\mathring{Y}$
that were already part of the computation of $\nabla_{H}$ (see Eq.
\ref{eq:gen-grad-H-fact} in section \ref{sub:gen-grad-H-fact}),
we can thus compute $M$ efficiently as

\begin{eqnarray}
M & = & 4\mathrm{diag}(\nabla_{q})\overbrace{H^{T}QH}^{\hat{M}}~\mathrm{diag}(\nabla_{q})+\left(D\nabla_{s}\nabla_{s}^{T}+\mathring{Y}^{T}\mathring{Y}+\nabla_{s}\mathbf{\bar{y}}^{T}+\mathbf{\bar{y}}\nabla_{s}^{T}\right)\nonumber \\
 &  & +2\left(\mathrm{diag}(\nabla_{q})H^{T}\hat{Z}\right)+2\left(\mathrm{diag}(\nabla_{q})H^{T}\hat{Z}\right)^{T}\label{eq:gen-M}
\end{eqnarray}

Note that computing $M$ requires computing $\mathring{Y}^{T}\mathring{Y}$,
a $m\times m$ matrix, each element of which is the dot product between
two $K-sparse$ columns of sparse matrix $\mathring{Y}$ so that it
can be computed in $O(m^{2}K)$.

Having $M$ we can then update $Q$ using Eq. \ref{eq:gen-Qnew}.

\subsection{Bookkeeping operations: tracking $U^{-T}$}

We can update$U^{-T}$ to reflect our rank-$m$ update of $U$ in
step a), using the Woodbury identity.

\subsection{Putting it all together}

In this section, we put together all the operations that we have derived
to write the minibatch version of the update algorithm for general
spherical losses.

The parameters of the output layer that we will learn are $V,U,\omega$
and implicitly represent $W$ as $W=VU+\mathbf{1}_{D}\omega^{T}$.

The algorithm will work for any spherical loss function $\ell$ in
canonical form that computes $\ell(q,s,\mathcal{K}_{,},\mathbf{a},\mathbf{t})$
and for which we can compute gradients with respect to its parameters.

\subsection*{Initialization}
\begin{itemize}
\item we can initialize $D\times d$ matrix $V$ randomly as we would have
initialized $W$ so that we initially have $V=W$.\\
Alternatively we can initialize $V$ to 0 (there won't be symmetry
breaking issues with having $W$ initially be 0 provided the other
layers are initialized randomly, since varying inputs and targets
will naturally break symmetry for the output layer)
\item we initialize $U$ to the identity: $U\leftarrow\mathbf{I}_{d}$ 
\item and $\omega$ to zero $\omega\leftarrow\mathbf{0}_{d}$ so that, trivially,
we initially have $VU+\mathbf{1}_{D}\omega^{T}=W$.
\item initialize $U^{-T}\leftarrow\mathbf{I}_{d}$
\item initialize $Q\leftarrow W^{T}W=V^{T}V$ (or more cheaply initialize
$Q\leftarrow0$ if we have initialized $V$ to 0).
\item initialize $\bar{w}=W^{T}\mathbf{1}_{D}=\mathrm{rowsum}(W)=\mathrm{rowsum}(V)$
(or more cheaply $\bar{w}\leftarrow0$ if we have initialized $V$
to 0).
\end{itemize}

\subsection*{Minibatch update algorithm for arbitrary spherical loss}

\textbf{Inputs} (besides above parameters $V,U,\omega$ and bookkeeping
variables $Q,U^{-T},\bar{w}$): 
\begin{itemize}
\item $H$ : a $d\times m$ matrix whose $m$ \uline{columns} contain
the last hidden layer representation vectors for $m$ example (with
an appended constant 1 element to account for an output bias). 
\item $Y$ : a $D\times m$ \uline{sparse} target matrix that uses sparse
representation $(\mathcal{K},T)$ so that $Y_{\mathcal{K}_{kj},j}=T_{kj}$
for $k\in\{1,\ldots,K\}$ and $j\in\{1,\ldots,m\}$. Each of the $m$
columns of $Y$ is the $K$-sparse target vector associated to one
example of the minibatch.
\item $\eta\in\mathbb{R}^{+}$learning rate for the update
\end{itemize}
\textbf{Updates:}
\begin{itemize}
\item parameters and bookkeeping matrices $U,V,\omega,Q,U^{-T},\bar{w}$
\end{itemize}
\textbf{Returns:} 
\begin{itemize}
\item $L\in\mathbb{R}$ the sum of squared error losses for the $m$ examples
of the minibatch
\item $\nabla_{H}$ a $d\times m$ matrix whose $m$ columns contain the
gradient of the loss with respect to $H$, to further backpropagate
upstream.
\end{itemize}
The detailed algorithm is given as Algorithm \ref{alg:minibatch-spherical-update}

\begin{algorithm}
\protect\caption{\label{alg:minibatch-spherical-update}Minibatch version of the update
algorithm for general spherical loss}

\textbf{FUNCTION} \textbf{spherical\_minibatch\_fbprop\_update:}

~~~Inputs: $\overbrace{H}^{\textrm{hidden layer minibatch}},\overbrace{\mathcal{K},T}^{\textrm{sparse target}},\overbrace{\eta}^{\textrm{learning rate}},~\overbrace{V,U,\omega}^{\textrm{layer parameters}},~\overbrace{Q,\bar{w},U^{-T}}^{\textrm{bookkeeping variables}}$

~~~Updates: $V,U,\omega,Q,\bar{w},U^{-T}$

~~~Returns: loss $L$, gradient $\nabla_{H}$ to backpropagate
further upstream 

\begin{tabular}{>{\raggedright}p{0.5\textwidth}>{\raggedright}p{0.1\textwidth}>{\raggedright}p{0.1\textwidth}>{\raggedright}p{0.1\textwidth}}
\textbf{\small{}Operations} & \textbf{\small{}main text Eq.}{\small{} } & \textbf{\small{}result dims} & \textbf{\small{}\# ops}\tabularnewline
{\small{}$\hat{H}=QH$} & {\small{}Eq. \ref{eq:qvec}} & {\small{}$d\times m$} & {\small{}$md^{2}$}\tabularnewline
{\small{}$\hat{M}=H^{T}\hat{H}$} & {\small{}Eq. \ref{eq:qvec}} & {\small{}$m\times m$} & \tabularnewline
{\small{}$\mathbf{q}=\mathrm{diag}(\hat{M}$)}{\small \par}

{\small{}} & {\small{}Eq. \ref{eq:qvec}} & {\small{}$m$} & {\small{}$m$}\tabularnewline
{\small{}$\mathbf{s}=H^{T}\bar{w}$} & {\small{}Eq. \ref{eq:svec}} & {\small{}$m$} & {\small{}$md$}\tabularnewline
{\small{}$\tilde{H}=UH$} & {\small{}Eq. \ref{eq:gen-Htilde}} & {\small{}$d\times m$} & {\small{}$md^{2}$}\tabularnewline
{\small{}$\tilde{\mathbf{h}}=H^{T}\omega$} & {\small{}Eq. \ref{eq:gen-htilde}} & {\small{}$m$} & {\small{}$md$}\tabularnewline
{\small{}}{\small \par}

{\small{}Matrix$A$: $A_{kj}=(\tilde{H}_{j})^{T}V_{\mathcal{K}_{kj}\bullet}+\tilde{\mathbf{h}}_{j}$} & {\small{}Eq. \ref{eq:gen-A-fact}} & {\small{}$K\times m$} & {\small{}$mKd$}\tabularnewline
{\small{}$\vec{L}=\left[\ell(\mathbf{q}_{j},\mathbf{s}_{j},\mathcal{K}_{j,},A_{j},T_{j})\right]_{j=1\ldots m}$} & {\small{}Eq. \ref{eq:Lvec}} & {\small{}$m$} & {\small{}typically $O(Km)$}\tabularnewline
{\small{}$L=\mathrm{sum}(\vec{L})$} &  & {\small{}1} & {\small{}$m$}\tabularnewline
{\small{}$\nabla_{q}=\left[\frac{\partial\ell}{\partial q}(\mathbf{q}_{j},\mathbf{s}_{j},\mathcal{K}_{j,},A_{j},T_{j})\right]_{j=1\ldots m}$} &  & {\small{}$m$} & \tabularnewline
{\small{}$\nabla_{s}=\left[\frac{\partial\ell}{\partial s}(\mathbf{q}_{j},\mathbf{s}_{j},\mathcal{K}_{j,},A_{j},T_{j})\right]_{j=1\ldots m}$} &  & {\small{}$m$} & \tabularnewline
{\small{}$\nabla_{A}=\left[\frac{\partial\ell}{\partial\mathbf{a}_{k}}(\mathbf{q}_{j},\mathbf{s}_{j},\mathcal{K}_{j,},A_{j},T_{j})\right]_{k=1\ldots K,~j=1\ldots m}$} &  & {\small{}$K\times m$} & \tabularnewline
{\small{}$\mathring{Y}=\mathrm{sparsemat}_{D,m}(\mathcal{K},\nabla_{A})$} &  & {\small{}$D\times m$ ($K$-sparse)} & \tabularnewline
{\small{}$\mathbf{\bar{y}}=\mathring{Y}^{T}\mathbf{1}_{D}=\mathrm{rowsum}(\nabla_{A})$} & {\small{}Eq. \ref{eq:ybar}} & {\small{}$m$} & {\small{}$Km$}\tabularnewline
{\small{}$\hat{Z}=\bar{w}\nabla_{s}^{T}+U^{T}(V^{T}\mathring{Y})+\omega\mathbf{\bar{y}}^{T}$} & {\small{}Eq. \ref{eq:gen-Zhat-fact}} & {\small{}$d\times m$} & {\small{}$md$}\tabularnewline
{\small{}$\nabla_{H}=2\hat{H}\mathrm{~diag}(\nabla_{q})+\hat{Z}$ } & {\small{}Eq. \ref{eq:gen-grad-H-fact}} & {\small{}$d\times m$} & {\small{}$md$}\tabularnewline
{\small{}$U\leftarrow U-2\eta(\underbrace{UH}_{\tilde{H}})~\mathrm{diag}(\nabla_{q})H^{T}$} & {\small{}Eq. \ref{eq:gen-updateU}} & {\small{}$d\times d$} & {\small{}$md^{2}$}\tabularnewline
{\small{}$U^{-T}\leftarrow...$ use Woodbury Identity to update it.} &  & {\small{}$d\times d$} & {\small{} $2m^{2}d+m^{3}+2md^{2}$}\tabularnewline
{\small{}$\omega\leftarrow\omega-\eta H(2\mathrm{diag}(\nabla_{q})\underbrace{H^{T}\omega}_{\tilde{\mathbf{h}}}+\nabla_{s})$} & {\small{}Eq. \ref{eq:gen-update-omega}} & {\small{}$d$} & {\small{}$2md+3d$}\tabularnewline
{\small{}$V\leftarrow V-\eta\mathring{Y}(U_{new}^{-T}H){}^{T}$} & {\small{}Eq. \ref{eq:gen-updateV}} & {\small{}$D\times d$} & {\small{}$md^{2}+mKd$}\tabularnewline
{\small{}$\bar{w}\leftarrow\bar{w}-\eta H\left(2\mathrm{diag}(\nabla_{q})H^{T}\bar{w}+D\nabla_{s}+\mathbf{\bar{y}}\right)$} & {\small{}Eq. \ref{eq:gen-update-wbar}} & {\small{}$d$} & {\small{}$2md+4d$}\tabularnewline
{\small{}$M=4\mathrm{diag}(\nabla_{q})\hat{M}~\mathrm{diag}(\nabla_{q})$}{\small \par}

{\small{}$~+D\nabla_{s}\nabla_{s}^{T}+\mathring{Y}^{T}\mathring{Y}+\nabla_{s}\mathbf{\bar{y}}^{T}+\mathbf{\bar{y}}\nabla_{s}^{T}$}\\
{\small{}$~+2\left(\mathrm{diag}(\nabla_{q})H^{T}\hat{Z}\right)+2\left(\mathrm{diag}(\nabla_{q})H^{T}\hat{Z}\right)^{T}$} & {\small{}Eq. \ref{eq:gen-M}} & {\small{}$m\times m$} & {\small{}}{\small \par}

{\small{}$2m^{2}d+(5+K)m^{2}+d^{2}$}\tabularnewline
{\small{}$Q\leftarrow Q-\eta\nabla_{H}H^{T}-\eta H\nabla_{H}^{T}+\eta^{2}(HM)H^{T}$} & {\small{}Eq. \ref{eq:gen-Qnew}} & {\small{}$d\times d$} & {\small{}$md^{2}+2m^{2}d+2d^{2}$}\tabularnewline
{\small{}RETURN $L$, $\nabla_{H}$} &  &  & \tabularnewline
\end{tabular}
\end{algorithm}

Counting the total number of basic operations of the update algorithm
yields roughly $8md^{2}+m^{3}+7m^{2}d+2mKd+3d^{2}\approx17md^{2}$
operations.

Comparing this$17md^{2}$ to the $3Dm$ of the naive update, the expected
theoretical speedup is approximately $\frac{3D}{18d}=\frac{1}{6}\frac{D}{d}$

For $d=512$ and $D=793471$ this yields a theoretical speedup of
258

Note that in the special cases where the specific loss function $\ell$
does not depend on the sum of outputs $s$ (as is the case e.g. of
the squared error) then we don't need to compute $s$, and can use
a$\omega$ that is always 0 so there's a lot we don't need to compute
and update.\\

\section{Controlling numerical stability}

\label{sec:stabilization}

The update of $U$ may over time lead to $U$ becoming ill-conditioned.
Simultaneously, as we update $U$ and $U^{-T}$ (using Sherman-Morrison
or Woodbury) our updated $U^{-T}$may over time start to diverge from
the true $U^{-T}$due to numerical precision. It is thus important
to prevent both of these form happening, i.e. make sure $U$ stays
well conditioned, to ensure the numerical stability of the algorithm.
We present here progressively refined strategies for achieving this.

\subsection{Restoring the system in a pristine stable state}

One simple way to ensure numerical stability is to once in a while
restore the system in its pristine state where $V=W$ and $U=\mathbf{I}_{d}=U^{-T}$.
This is easily achieved as follows:

\begin{eqnarray*}
V & \leftarrow & VU\\
U & \leftarrow & \mathbf{I}_{d}\\
U^{-T} & \leftarrow & \mathbf{I}_{d}.
\end{eqnarray*}
This operation doesn't affects the product $VU$, so the implicit
matrix $W$ remains unchanged, nor does it affect $Q=W^{T}W$. And
it does restore $U$ to a perfectly well conditioned identity matrix.
But computing $VU$ is an extremely costly $O(Dd^{2})$ operation,
so if possible we want to avoid it (except maybe once at the very
end of training, if we want to compute the actual $W$). In the next
paragraphs we develop a more efficient strategy.

\subsection{Stabilizing only problematic singular values}

$U$ becoming ill-conditioned is due to its singular values over time
becoming too large and/or too small. Let use define $\sigma_{1},~\ldots,~\sigma_{d}$
as the singular values of $U$ ordered in decreasing order. The conditioning
number of $U$ is defined as $\frac{\sigma_{1}}{\sigma_{d}}$ and
it can become overly large when $\sigma_{1}$ becomes too large and/or
when $\sigma_{d}$ becomes too small. Restoring the system in its
pristine state, as shown in the previous paragraph, in effect brings
back \emph{all} singular values of $U$ back to 1 (since it brings
back $U$ to being the identity). It is instead possible, and computationally
far less costly, to correct when needed only for the singular values
of $U$ that fall outside a safe range. Most often we will only need
to occasionally correct for one singular value (usually the smallest,
and only when it becomes too small). Once we have determined the offending
singular value and its corresponding singular vectors, correcting
for that singular value, i.e. effectively bringing it back to 1, will
be a $O(Dd)$ operation. The point is to apply corrective steps only
on the problematic singular values and only when needed, rather than
blindly, needlessly and inefficiently correcting for all of them through
the basic $O(Dd^{2})$ full restoration explained in the previous
paragraph.

Here is the detailed algorithm that achieves this:

\begin{algorithm}[H]
\protect\caption{Numerical stabilization procedure for problematic singular values}

\begin{itemize}
\item The chosen safe range for singular values is $[\sigma_{\mathrm{low}}, \sigma_{\mathrm{high}}]$ (ex: $[0.001, 100]$ )
\item The procedures given below act on output layer parameters $U$, $U^{-T}$ and $V$. 
\item For concision, we do not enlist these parameters explicitly in their parameter list.
\item Procedure \textsc{singular-stabilize} gets called after every $n_{\mathrm{check}}$ gradient updates (ex: $n_{\mathrm{check}}=100$).
\end{itemize}

\begin{algorithmic} 

\Procedure{singular-stabilize}{$\,$}
  \State $\bar{\mathbf{U}}$, $\sigma$, $\bar{\mathbf{V}}$ = \Call{SVD}{$U$} \Comment{Computes singular value decomposition of $U$ as $U = \bar{\mathbf{U}}\, \mathrm{diag}(\sigma)\, \bar{\mathbf{V}}^T $}
  \ForAll{ $k \in \{1, \ldots, d \}$ }
    \If{$\sigma_k$ < $\sigma_{\mathrm{low}}$ OR $\sigma_k$ > $\sigma_{\mathrm{high}}$}
      \State \Call{fix-singular-value}{$\sigma_k$, $\bar{\mathbf{U}}_k$, $1$}
    \EndIf
  \EndFor
\EndProcedure

\vspace*{0.5cm}

\noindent \emph{The following procedure will change singular value $\sigma$ of $U$ associated to singular vector $u$ to become target singular value $\sigma^*$ (typically 1). It doesn't change $U$'s singular vectors, only that one singular value. It also changes $V$ symetrically (with a rank-one update) in such a way that $W=VU$ remains unchanged.}
\vspace*{0.5cm}

\Procedure{fix-singular-value}{$\sigma$, $u$, $\sigma^*$} 
\State $ \alpha = \frac{\sigma^*-\sigma}{\sigma}$
\State $ \beta = - \frac{\alpha}{1+\alpha}$
\State $U \leftarrow U + \alpha u (U^T u)^T$
\State $V \leftarrow V + \beta (V u) u^T $
\State $U^{-T} \leftarrow U^{-T} + \beta u (U^{-1} u)^T$ \Comment{Where $U^{-1}$ is obtained as the transpose of $U^{-T}$. But we may instead of this prefer to recompute $U^{-T}$ from scratch by inverting $U$ to ensure it doesn't stray too much due to numerical imprecisions.}
\EndProcedure

\end{algorithmic}
\end{algorithm}

Proof that $W=VU$ is left unchanged by \noun{fix-singular-value}

\begin{eqnarray*}
V_{new}U_{new} & = & (V+\beta(Vu)u^{T})~(U+\alpha u(U^{T}u)^{T})\\
 & = & V(\mathbf{I}_{d}+\beta uu^{T})~(U+\alpha uu^{T}U)\\
 & = & V(\mathbf{I}_{d}+\beta uu^{T})~(\mathbf{I}_{d}+\alpha uu^{T})U\\
 & = & V(\mathbf{I}_{d}^{2}+\beta uu^{T}+\alpha uu^{T}+\beta\alpha uu^{T}uu^{T})U\\
 & = & V(\mathbf{I}_{d}^{2}+(\alpha+\beta)uu^{T}+\beta\alpha u(u^{T}u)u^{T})U\\
 & = & V(\mathbf{I}_{d}+(\alpha+\beta)uu^{T}+\beta\alpha uu^{T})U\\
 & = & V(\mathbf{I}_{d}+(\alpha-\frac{\alpha}{1+\alpha}+\alpha\frac{-\alpha}{1+\alpha})uu^{T})U\\
 & = & V(\mathbf{I}_{d}+(\alpha-\frac{\alpha}{1+\alpha}-\frac{\alpha^{2}}{1+\alpha})uu^{T})U\\
 & = & V(\mathbf{I}_{d}+(\alpha-\frac{\alpha+\alpha^{2}}{1+\alpha})uu^{T})U\\
 & = & V(\mathbf{I}_{d}+(\alpha-\frac{\alpha(1+\alpha)}{1+\alpha})uu^{T})U\\
 & = & V(\mathbf{I}_{d}+(\alpha-\alpha)uu^{T})U\\
 & = & V\mathbf{I}_{d}U\\
 & = & VU
\end{eqnarray*}

\subsection{Avoiding the cost of a full singular-value decomposition}

Computing the SVD of $d\times d$ matrix $U$ as required above, costs
roughly $25d^{3}$ elementary operations (use the so-called \noun{r-svd}
algorithm). But since the offending singular values will typically
be only the smallest or the largest, it is wasteful to compute all
$d$ singular values every time. A possibly cheaper alternative is
to use the power iteration method with $U$ to find its largest singular
value and associated singular vector, and similarly with $U^{-1}$to
obtain the smallest singular value of $U$ (which corresponds to the
inverse of the largest singular value of $U^{-1}$). Each iteration
of the power iteration method requires only $O(d^{2})$ operations,
and a few iterations may suffice. In our experiments we fixed it to
100 power iterations. Also it is probably not critical if the power
iteration method is not run fully to convergence, as correcting along
an approximate offending singular vector direction may be sufficient
for the purpose of ensuring numerical stability. 

With this refinement, we loop over finding the smallest singular value
with the power iteration method, correcting for it to be 1 by calling
\noun{fix-singular-value} if it is too small, and we repeat this until
we find the now smallest singular value to be inside the acceptable
range. Similarly for the largest singular values. 

Note that while in principle we may not need to ever invert $U$ from
scratch (as we provided update formulas of $U^{-T}$ with every change
we make to $U$), it nevertheless proved to be necessary to do so
regularly to ensure $U^{-T}$ doesn't stray too much from the correct
value due to numerical imprecisions. Inverting $U$ using Gaussian-elimination
costs roughly $d^{3}$ operations, so it is very reasonable and won't
affect the computational complexity if we do it no more often than
every $d$ training examples (which will typically correspond to less
than 10 minibatches of size 128). In practice, we recompute $U^{-T}$
from scratch every time before we run this check for singular value
stabilization.

\section{Experimental validation}

\label{sec:experiments}

\begin{figure}
\includegraphics[width=0.5\textwidth]{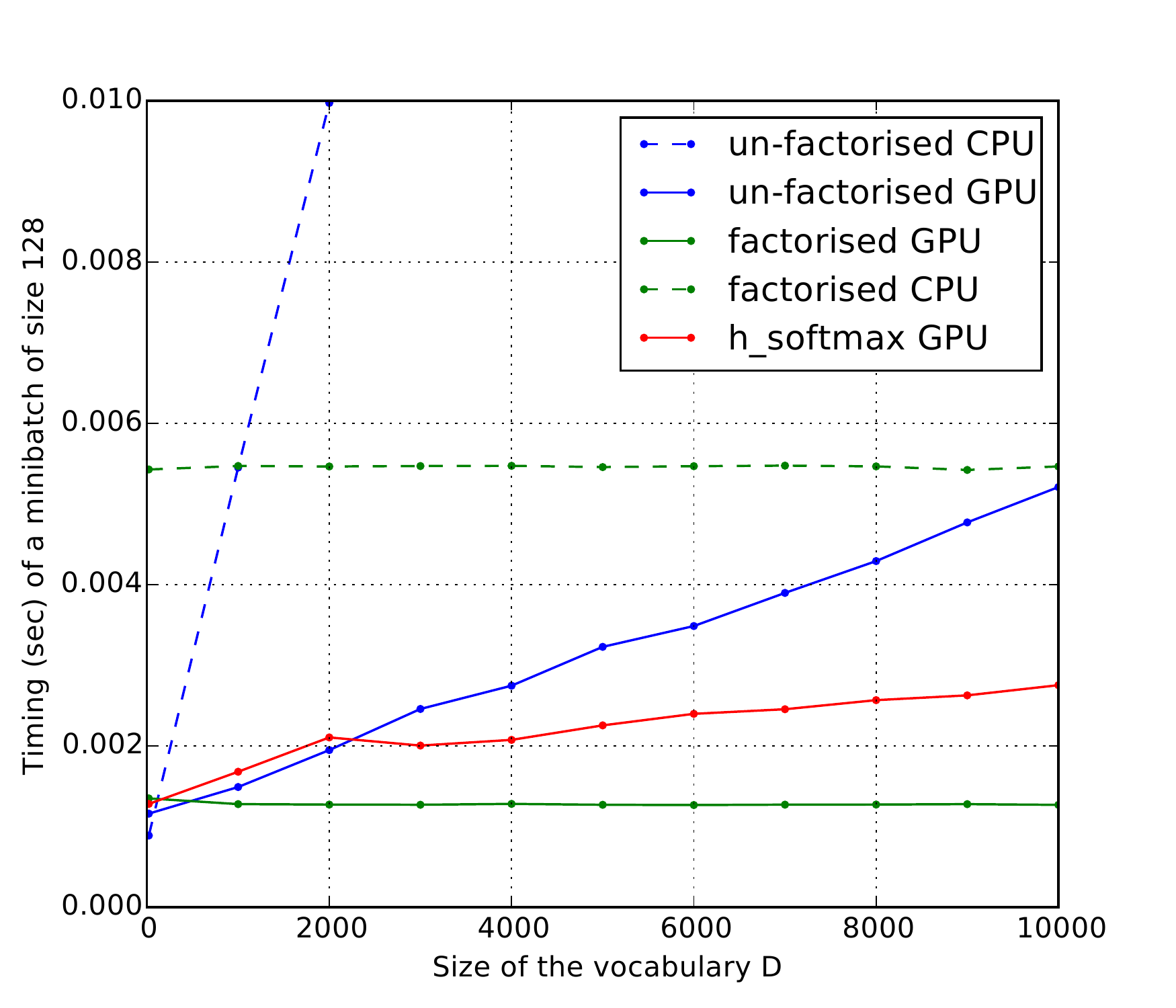}\includegraphics[width=0.5\textwidth]{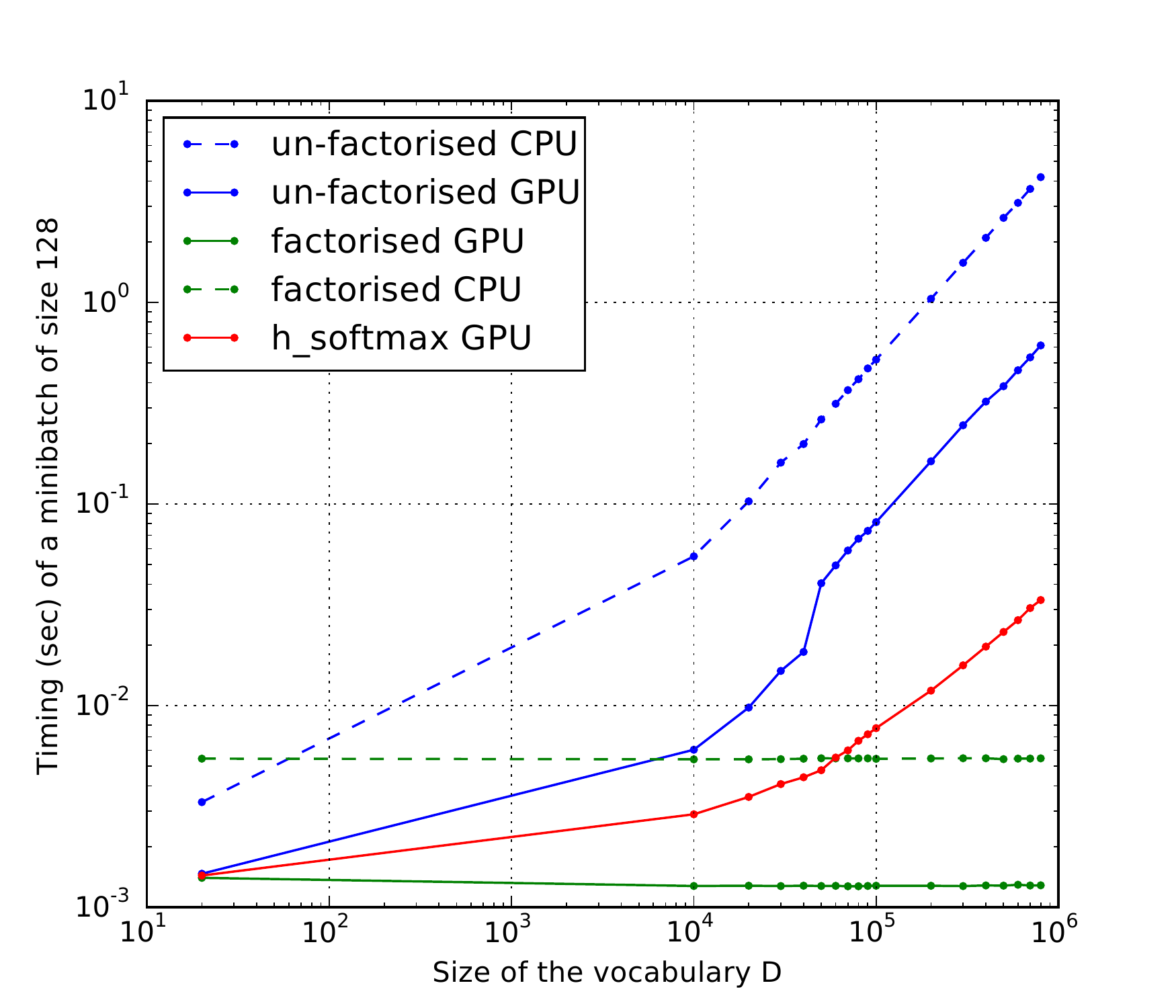}

\protect\caption{\label{fig:Timing-of-different}Timing of different algorithms. Time
taken by forward and backward propagations in the output layer, including
weight update, on a minibatch of size 128 for different sizes of vocabulary
D on both CPU and GPU. The input size d is fixed to 300. The Timing
of a 2 layer hierarchical softmax efficient GPU implementation (h\_softmax)
is also provided for comparison. Right plot is in log-log scale. As
expected, the timings of factorized versions are independent of the
size of the vocabulary.}

\end{figure}

\begin{figure}
\includegraphics[width=0.5\textwidth]{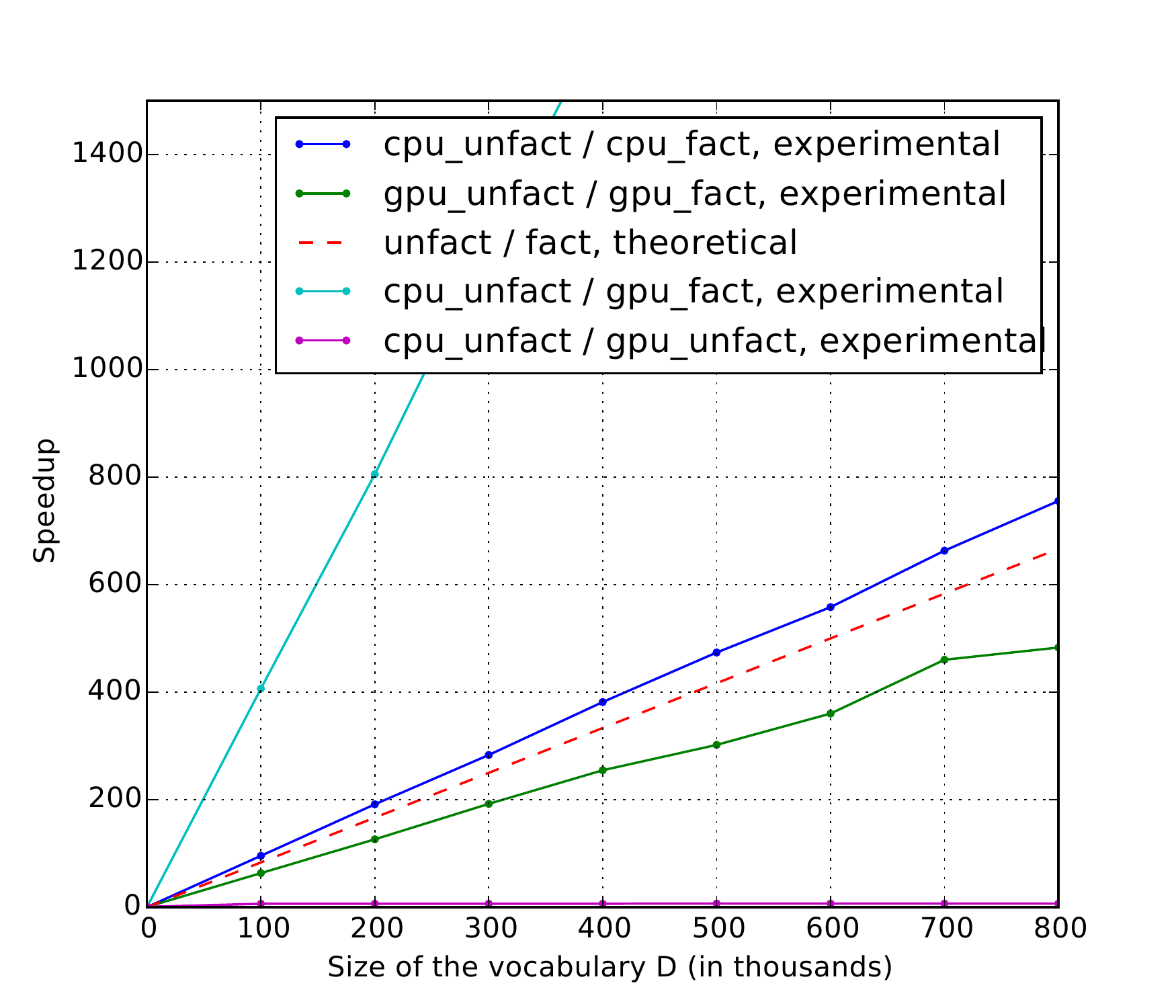}\includegraphics[width=0.5\textwidth]{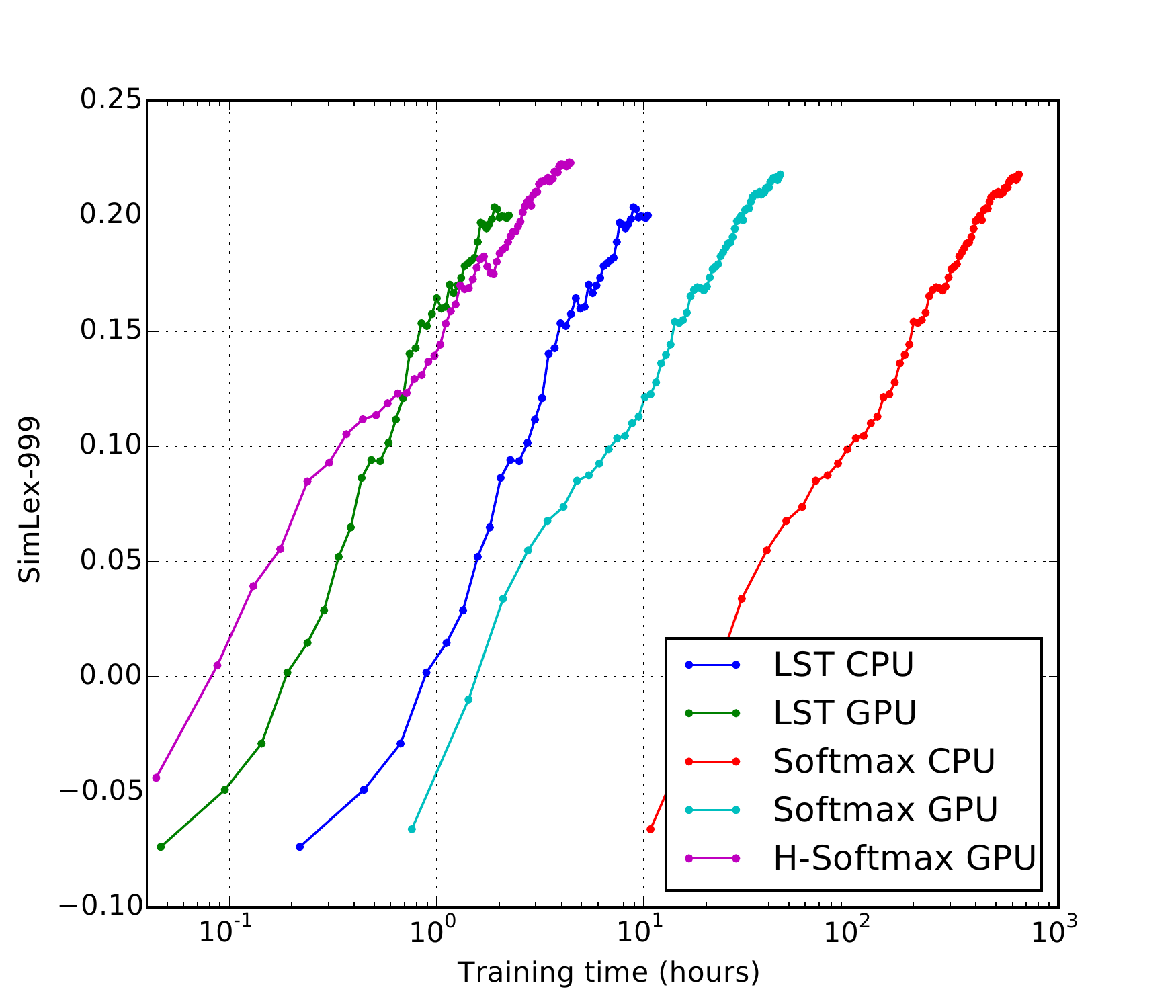}\protect\caption{\textbf{\label{fig:speedup}Left:} Practical and theoretical speedups
for different sizes of vocabulary D and fixed input size d=300. The
practical unfact / fact speedup is similar to the theoretical one.
\textbf{Right}: Evolution of the Simlex-999 score obtained with different
models as a function of training time (CPU softmax times were extrapolated
from fewer iterations). Softmax models are zero hidden-layer models,
while our large sparse target (LST) models have two hidden layers.
These were the best architectures retained in both cases (surprisingly
the softmax models with hidden layers performed no better on this
task). The extra non-linear layers in LST may help compensate for
the lack of a softmax. LST models converge to slightly lower scores
at similar speed as the hierarchical softmax model but significantly
faster than softmax models. }
\end{figure}

We implemented both a CPU version using \emph{blas} and a parallel
GPU (Cuda) version using \emph{cublas} of the proposed algorithm\footnote{Open source code will be released upon official publication of this
research.}. We evaluated the GPU and CPU implementations by training word embeddings
with simple neural language models, in which a probability map of
the next word given its preceding n-gram is learned by a neural network.
We used a Nvidia Titan Black GPU and a i7-4820K @ 3.70GHz CPU and
ran experiments on the one billion word dataset\citep{DBLP:conf/interspeech/ChelbaMSGBKR14},
which is composed of 0.8 billions words belonging to a vocabulary
of 0.8 millions words. We evaluated the resulting word embeddings
with the recently introduced Simlex-999 score \citep{DBLP:journals/corr/HillRK14},
which measures the similarity between words. We also compared our
approach to unfactorised versions and to a two-layer hierarchical
softmax. Figure \ref{fig:Timing-of-different} and \ref{fig:speedup}
(left) illustrate the practical speedup of our approach for the output
layer only. Figure \ref{fig:speedup}(right) shows that the LST (Large
Sparse Target) models are much faster to train than the softmax models
and converge to only slightly lower Simlex-999 scores. Table \ref{tab:Speedups-with-respect}
summarizes the speedups for the different output layers we tried,
both on CPU and GPU. We also emprically verified that our proposed
factored algorithm learns the exact same model weights $(VU)$ as
the corresponding naive unfactored algorithm's $W$, as it theoretically
should (up to negligible numerical precision differences), and followed
the exact same learning curves (as a function of number of iterations,
not time!). 

\begin{table}
\centering{}\protect\caption{\label{tab:Speedups-with-respect}Speedups with respect to the baseline
naive model on CPU, for a minibatch of 128 and the whole vocabulary
of D = 793471 words. This is a two hidden layer model with 300 neurons
on all its layers (so d = 300). }
\begin{tabular}{|c|c|c|}
\hline 
\textbf{Model} & \textbf{output layer only speedup} & \textbf{whole model speedup}\tabularnewline
\hline 
\hline 
cpu unfactorised (naive) & 1 & 1\tabularnewline
\hline 
gpu unfactorised (naive) & 6.8 & 4.7\tabularnewline
\hline 
gpu hierarchical softmax & 125.2 & 178.1\tabularnewline
\hline 
cpu factorised & 763.3 & 501\tabularnewline
\hline 
gpu factorised & 3257.3 & 1852.3\tabularnewline
\hline 
\end{tabular}
\end{table}

\section{Conclusion and future work}

We introduced a new algorithmic approach to efficiently compute the
\emph{exact} gradient updates for training deep networks with very
large sparse targets. Remarkably the complexity of the algorithm is
independent of the target size, which allows tackling very large problems.
Our CPU and GPU implementation yield similar speedups to the theoretical
one and can thus be used in practical applications, which could be
explored in further work. In particular, neural language models seem
good candidates. But it remains unclear how using a loss function
other than \emph{log-softmax} may affect the quality of the resulting
word embeddings and further research should be carried out in this
direction. While restricted, the spherical family of loss functions,
offers opportunities to explore alternatives to the ubiquitous softmax,
that thanks to the algorithm presented here, could scale computationally
to extremely large output spaces.

\section*{Acknowledgements}

We would like to thank the developers of Theano \citep{bergstra+al:2010-scipy,Bastien-Theano-2012}
and Blocks \citep{blocksfuel}. 

This research is supported by NSERC and Ubisoft.

\small

\bibliographystyle{unsrtnat_without_url}
\bibliography{strings,strings-shorter,ml,aigaion-shorter,large_sparse}

\normalsize
\end{document}